\documentclass{article}
\usepackage{arxiv}
\usepackage[utf8]{inputenc} % allow utf-8 input
\usepackage[T1]{fontenc}    % use 8-bit T1 fonts
\usepackage{hyperref}       % hyperlinks
\usepackage{url}            % simple URL typesetting
\usepackage{booktabs}       % professional-quality tables
\usepackage{nicefrac}       % compact symbols for 1/2, etc.
\usepackage{microtype}      % microtypography
\usepackage{cite} % Tidies up citation numbers.
\usepackage{graphicx}
\usepackage{tabularx}
\usepackage{lipsum}% Just for this example
\usepackage{caption}
\usepackage{multicol}
\usepackage{tikz}
\usetikzlibrary{positioning}% To get more advances positioning options
\usetikzlibrary{arrows}
\usepackage{bbm}
\usepackage{amsmath}
\usepackage{amsmath,amssymb}

\usepackage{bm}
\usepackage{subcaption} 
\usepackage{pgf}
\usepackage{epsfig}
\usepackage[figuresright]{rotating}
\usepackage{geometry}
\usepackage{amsmath,amsfonts,amssymb,amsthm}
\usepackage{mathtools}
\usepackage{commath}
\usepackage[sc,osf]{mathpazo}
\usepackage[font=footnotesize,labelfont=bf]{caption}
\usepackage[]{algorithm2e}
\usepackage{algorithmicx}
\usepackage[belowskip=0.0pt,aboveskip=0.0pt]{caption} 
\usepackage{amssymb}
\usepackage{tikz}

\usepackage{multirow}

\usepackage{hyperref}
\usepackage{lscape}
\usepackage{cleveref}

\title{Learning functionals via LSTM neural networks for predicting vessel dynamics in extreme sea states}
\author{
  Jos\'e del \'Aguila Ferrandis\thanks{Personal webpage: jaguila.mit.edu} \\ %---\emph{not} for acknowledging funding agencies
  Department of Mechanical Engineering\\
  Massachusetts Institute of Technology\\
  \texttt{jaguila@mit.edu} \\
  \And
  Michael Triantafyllou \\
  Department of Mechanical Engineering\\
  Massachusetts Institute of Technology\\
  \texttt{mistetri@mit.edu} \\
  \And
  Chryssostomos Chryssostomidis \\ %---\emph{not} for acknowledging funding agencies
  Department of Mechanical Engineering\\
  Massachusetts Institute of Technology\\
  \texttt{chrys@mit.edu}\\
   \And  
  George Karniadakis \\ %---\emph{not} for acknowledging funding agencies
  Division of Applied Mathematics,\\
  Brown University\\
  \texttt{george\_karniadakis@brown.edu}\\
}

\begin{document}

%%%% Article title to be placed here
\maketitle

%%%% Keyword entries to be placed here %%%%
%\keywords{LSTM Neural Networks, Extreme sea states, Seakeeping, Nonlinear functionals}

%%%% Abstract text to be placed here %%%%%%%%%%%%
\begin{abstract}
Predicting motions of vessels in extreme sea states represents one of the most challenging problems in naval hydrodynamics. It involves computing complex nonlinear wave-body interactions, hence taxing heavily computational resources.  Here, we put forward a new simulation paradigm by training recurrent type neural networks (RNNs) that take as input the stochastic wave elevation at a certain sea state and output the main vessel motions, e.g., pitch, heave and roll. We first compare the performance of standard RNNs versus GRU and LSTM neural networks (NNs) and show that LSTM NNs lead to the best performance. We then examine the testing error of two representative vessels, a catamaran in sea state 1 and a battleship in sea state 8. We demonstrate that good accuracy is achieved for both cases in predicting the vessel motions for unseen wave elevations. We train the NNs with expensive CFD simulations {\em offline}, but upon training, the prediction of the vessel dynamics {\em online} can be obtained at a fraction of a second. This work is motivated by the universal approximation theorem for functionals~\cite{Functionals}, and it is the first implementation of such theory to realistic engineering problems.
\end{abstract}
%%%%%%%%%%%%%%%%%%%%%%%%%%%

%%%%%%%%%% Insert the texts which can accomdate on firstpage in the tag "fmtext" %%%%%

%%%%%%%%%%%%%%% End of first page %%%%%%%%%%%%%%%%%%%%%

\section{Brief Discussion of the Physical Problem}
\label{sec:cfd}

Predicting the motion of vessels in nonlinear waves constitutes one of the most challenging problems in naval hydrodynamics. A complete solution of the seakeeping problem involves resolving complex nonlinear wave-body interactions that may require  computational times of 100s of hours on multi-processor computers. For this reason, over the past few decades different seakeeping models have been formulated in order to predict vessel motions using simplified flow theories \cite{lee2005computation,newman1988computation,newman1977marine,newman1985algorithms,newman1986distributions}. These numerical methods are usually based on potential flow theories solved within the linear assumption, which limits their range of applicability and are inaccurate in severe sea states. Furthermore, second-order potential-flow solvers \cite{faltinsen1993sea,Faltinsen1988,Faltinsen2015,Faltinsen1990,faltinsen1987slow,Kring1994,Kring1999,Kring2000,Kring2004,beck1995desingularized,cao1994nonlinear} represent the state-of-the-art numerical methods for wave-induced response of vessels and other large-volume marine structures. Nevertheless, important viscous damping effects, such as those required in modeling of rolling of ships and slow-drift motions of moored structures, are not numerically resolved and  are accounted for by empirical formulas. 

\begin{figure}[h!] 
	\centering
    \begin{subfigure}[t]{0.49\textwidth}
        \centering
        \includegraphics[width=\columnwidth]{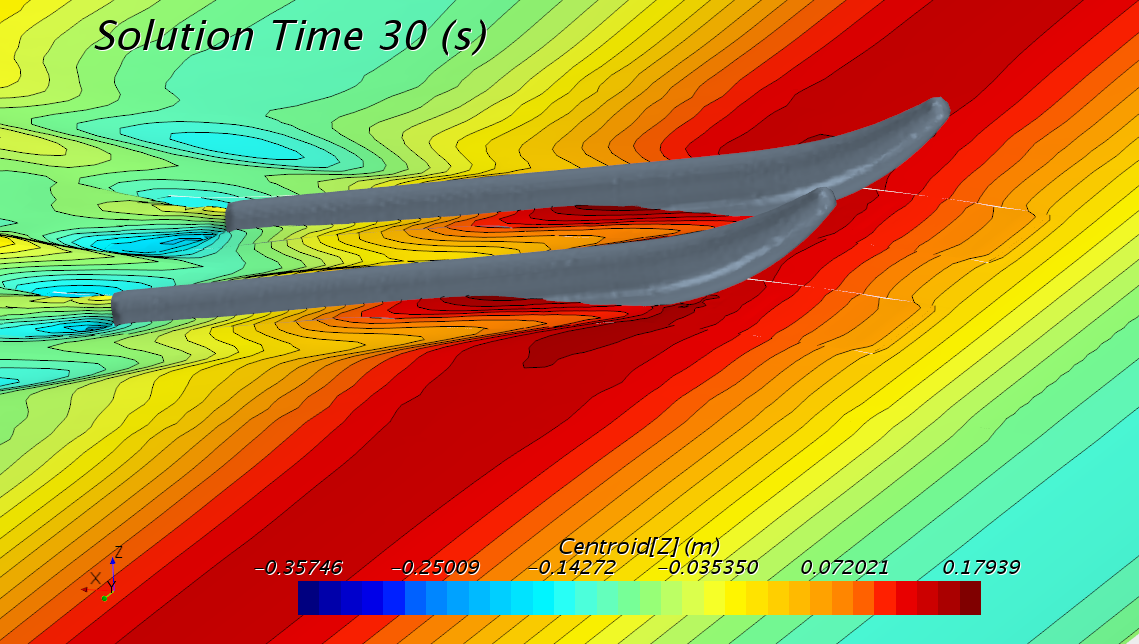}
        \caption{\scriptsize}
        \label{fig:Catamaran}
    \end{subfigure}%
~
    \begin{subfigure}[t]{0.49\textwidth}
        \centering
        \includegraphics[width=\columnwidth]{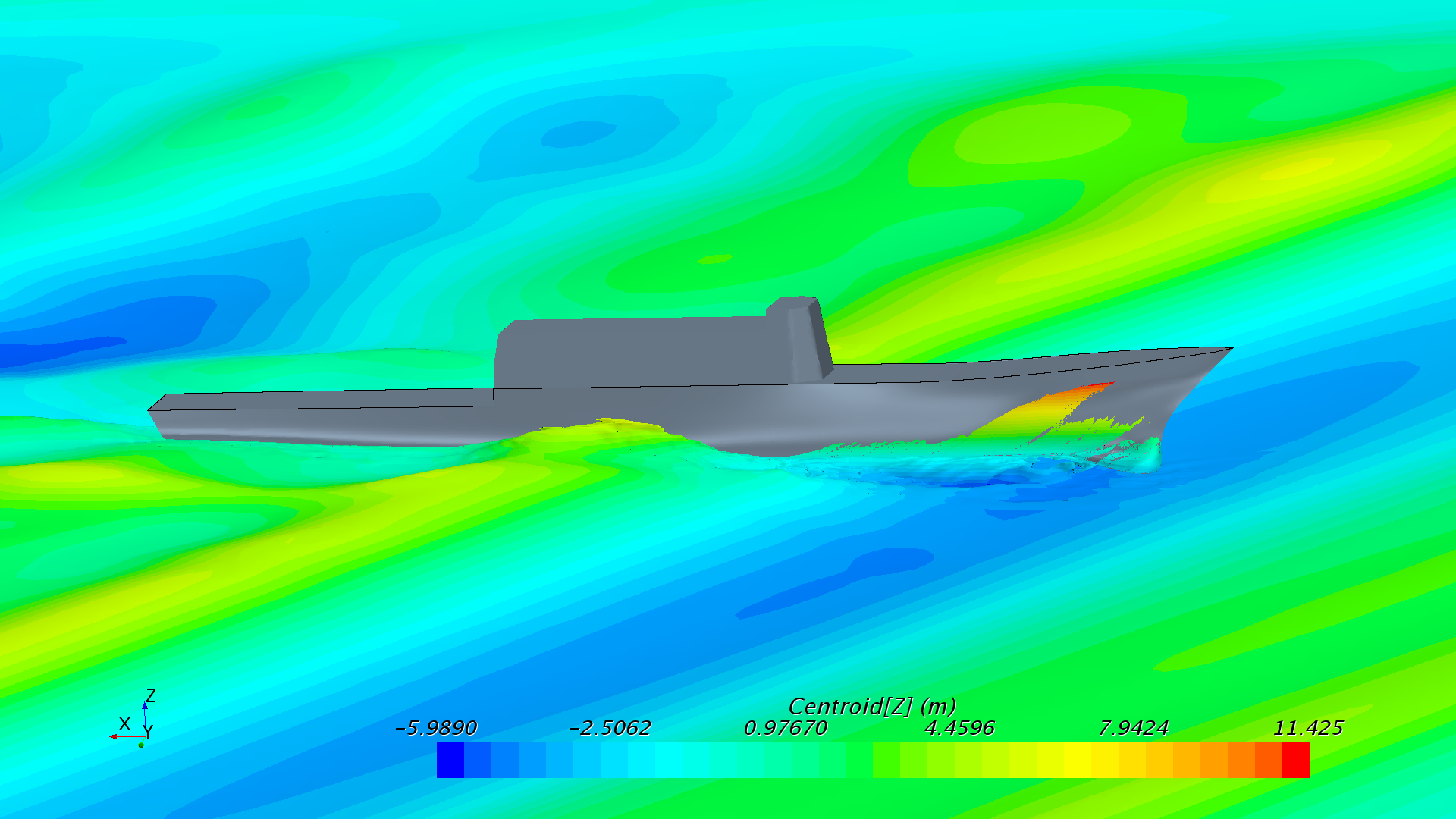}
        \caption{\scriptsize}
        \label{fig:DTMB}
    \end{subfigure}%
    \medskip    
    \caption{\textit{Snapshots of the URANS simulations}; the color scale (meters) represents sea surface elevation. (a) Catamaran sailing in regular 5th-order Stokes Waves. This constitutes a relatively easy condition to simulate using potential flow methods. (b) Notional DTMB battleship sailing in World Meteorological Organization (WMO) sea state 8 at Froude number 0.4. Meshes for this case involve more than 20 million finite volume cells and require several days to compute on a parallel computer with 300 cores. (See video for DTMB sailing through sea state 8 in supplementary materials).}
\label{fig:Sim}
\end{figure}

Disregarding viscous effects \textcolor{black}{might largely underestimate} the energy dissipated by the system while moving at the free surface, especially when analyzing ship rolling. This problem is particularly relevant for unconventional floating bodies at resonance, when waves are steep and nonlinear and when the motions of the vessels are large, compared it its main dimensions.  For this reason,  here we use a fully viscous model in the form of a Unsteady Reynolds Averaged Navier-Stokes (URANS) solver  \cite{ferziger2012computationa,itemreference3,ruth2015simulation,shen2013rans,Stern1,Stern2,menter1994two,VOF} to jointly resolve the viscous and nonlinear effects so that they can be learned by the RNN. A drawback is that URANS depends on empirical modelling of turbulence. Motion predictions obtained using fully viscous models are limited by the massive computational costs required. Among other applications, the approach taken here can be aimed at reducing the amount of simulation that is necessary to characterize motions in the operating conditions that the vessel/platform will encounter during its lifespan.

The data set is obtained from a viscous Volume-of-Fluid (VOF) URANS solver (STAR-CCM+). The equations solved are the averaged continuity and momentum equations for incompressible fluids:

\small
%%%%%%%%%%%%%%%%%%%%%%%%%%%%%%%%%%%%%%%%%%%%%%%%%%%%%%%%%%%%%%%%%%%%%%%%%%%%%%%%%%%%%%%%%%%%%%%%%
\begin{equation}
\frac{\partial (\rho \overline{u}_{i})}{\partial x_{i}} = 0 ,
\label{ACME 1}
\end{equation}
%%%%%%%%%%%%%%%%%%%%%%%%%%%%%%%%%%%%%%%%%%%%%%%%%%%%%%%%%%%%%%%%%%%%%%%%%%%%%%%%%%%%%%%%%%%%%%%%%
\begin{equation}
\frac{\partial (\rho \overline{u}_{i})}{\partial t} + 
\frac{\partial}{\partial x_{j}}(\rho \overline{u}_{i}\overline{u}_{j}+
\rho \overline{u_{i}'u_{j}'}) =
\frac{\partial \overline{p}}{\partial x_{i}}+\frac{\partial \overline{\tau}_{ij}}{\partial x_{j}} ,
\label{ACME 2}
\end{equation}
%%%%%%%%%%%%%%%%%%%%%%%%%%%%%%%%%%%%%%%%%%%%%%%%%%%%%%%%%%%%%%%%%%%%%%%%%%%%%%%%%%%%%%%%%%%%%%%%%
\begin{equation}
\overline{\tau}_{ij}=\mu\bigg(\frac{\partial \overline{u}_{i}}{\partial x_{j}}+\frac{\partial \overline{u}_{j}}{\partial x_{i}}\bigg),
\label{ACME 3}
\end{equation}
%%%%%%%%%%%%%%%%%%%%%%%%%%%%%%%%%%%%%%%%%%%%%%%%%%%%%%%%%%%%%%%%%%%%%%%%%%%%%%%%%%%%%%%%%%%%%%%%%
\normalsize

where $\overline{\tau}_{ij}$, in \cref{ACME 2}, are the components of the averaged viscous force tensor, $\overline{p}$ is the averaged pressure and $\overline{u}$ are the Cartesian components of the averaged velocity. In \cref{ACME 2}, $\overline{u_{i}'u_{j}'}$  are the Reynolds stresses, $\rho$ the fluid density, and $\mu$ the dynamic viscosity.

To model the free surface, the time-domain fully viscous model uses a VOF method~\cite{menter1994two}. This model assumes that the same equations governing the physics of one of the phases can be employed for all phases present in the computational domain (each cell or finite volume). A good reference for the theory behind this type of numerical method can be found in \cite{ferziger2012computationa}. %It assumes that these phases will have the same velocity, pressure and temperature.

In order to simulate the dynamic behavior and to obtain realistic platform motions, a Dynamic Fluid Body Interaction (DFBI) model is used. The vessel is allowed to move in two (catamaran vessel, see \cref{fig:Catamaran}) or three (DTMB vessel, see \cref{fig:DTMB}) degrees of freedom (DOF). Specifically, it is allowed to translate in the vertical direction (heave), to rotate around the transversal direction (pitch) and to rotate around the longitudinal direction (roll).

\begin{figure}[h!] 
    \centering
        \includegraphics[width=0.65\columnwidth]{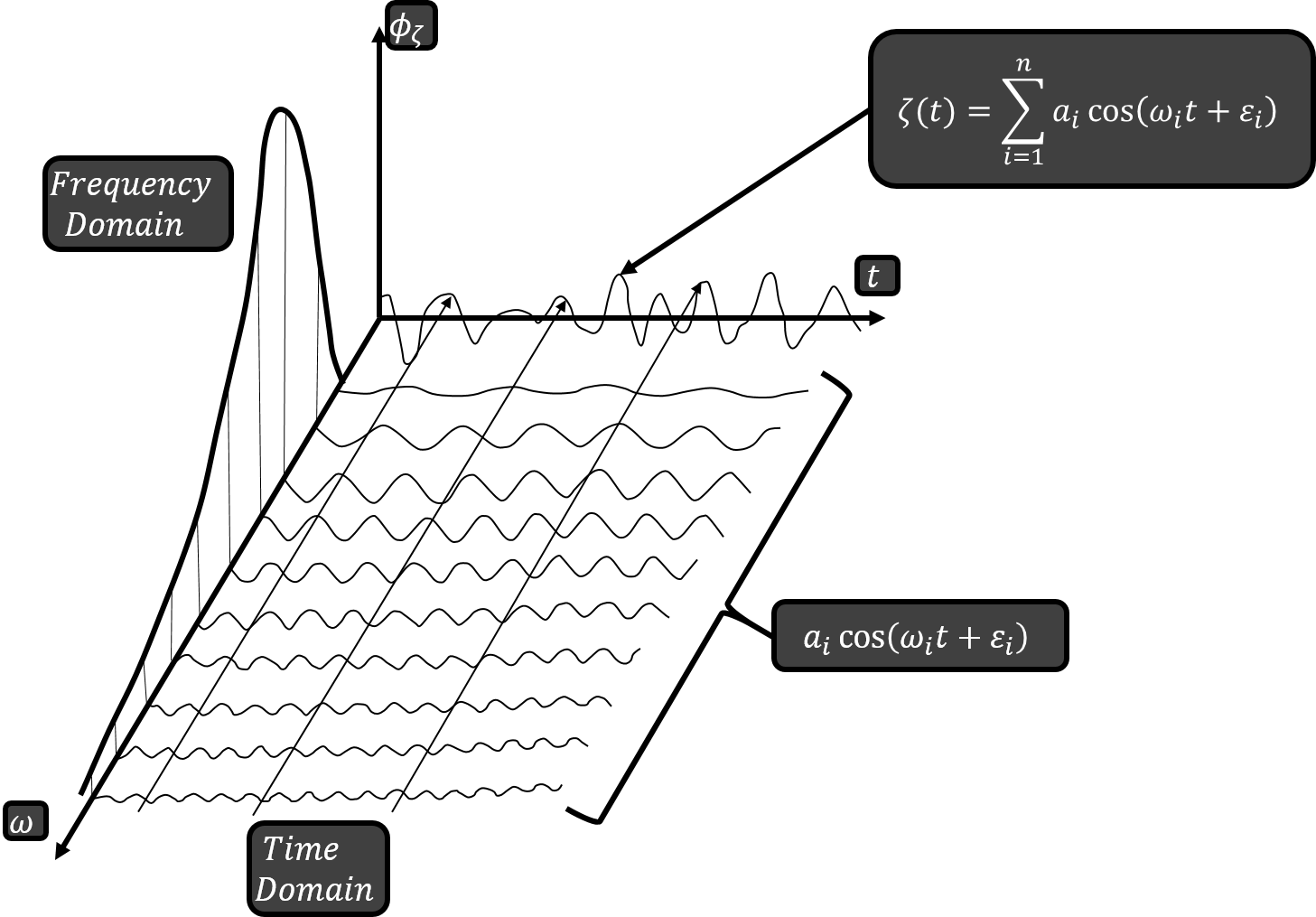}
    \medskip
    \caption{\textit{Graphical description of the spectral analysis performed to characterize sea states.} Given an energy spectrum specific to a particular ocean region, we can decompose the waves into different frequencies. The amplitudes corresponding to each frequency are given by the energy contained in each frequency in the wave spectra. Also, $\epsilon_i$ are random phases sampled from a uniform distribution.}
\label{fig:IrregSeasGraphics}
\end{figure}  

\subsection{Simulation of Irregular Long Crested Seas}

We are able to describe the waves that develop at the sea surface in a probabilistic way assuming it is stationary, homogeneous and ergodic random process. Then we represent the wave elevation through a finite sum of individual components approximating a Stieltjes integral, as found in \cite{schueller2012stochastic,shinozuka1987digital}.

Ocean waves used to induce motions in the vessels are reconstructed from experimental sea spectra (see \cref{fig:IrregSeasGraphics}) that characterize the stochastic process of sea surface elevations. At a particular spatial location, let $\zeta(t)$ be the sea surface elevation as a function of time. Then, this time signal can be defined as a sum of sinusoidal waves with random phases $\varepsilon_{i}$ between $-\pi$ and $\pi$ sampled from a uniform distribution, and incommensurate frequencies $\omega_i$ spanning the frequency range of the spectrum, i.e., 

\begin{equation}
   \zeta(t)=\sum_{i=1}^{n} a_{i} \cos \left(\omega_{i} t+\varepsilon_{i}\right) ,  
\end{equation}

which is a Gaussian probability density function as $n$ becomes large in accordance with the central limit theorem. The wave amplitude for a given frequency is obtained from the following relation,

\begin{equation}
    \frac{1}{2} a_{i}^{2} \cong S\left(\omega_{i}\right) \Delta \omega ,
\end{equation}

where $S\left(\omega\right)$ is the modified Pierson-Moskowitz spectrum~\cite{PMS} with $T_1$ as the mean wave period:

\begin{equation}
    \frac{S(\omega)}{H_s^{2} T_{1}}=\frac{0.11}{2 \pi}\left(\frac{\omega T_{1}}{2 \pi}\right)^{-5} \exp \left[-0.44\left(\frac{\omega T_{1}}{2 \pi}\right)^{-4}\right] ,
\end{equation}

 where 

\begin{equation}
    Y=\exp \left[-\left(\frac{0.191 \omega T_{1}-1}{2^{1 / 2} \sigma}\right)^{2}\right] ,
\end{equation}

and

\begin{equation}
    \sigma=\left\{\begin{array}{ll}{0.07} & {\text { if } \omega \leq 5.24 / T_{1}} \\ {0.09} & {\text { if } \omega>5.24 / T_{1}}\end{array}\right.   
    .
\end{equation}

To impose the initial conditions in the domain the velocity and pressure fields are calculated as a superposition of the individual regular waves. In  \cref{tab:irr} we present the general formulation for long crested irregular seas, where both space ($x_1,x_2$) and time ($t$) are independent variables. These are integrated numerically in time, also tracking nonlinear interactions with the vessels.

The sea states modeled vary for each of the two vessels considered here. In both cases we use the Pierson-Moskowitz spectrum to generate fully developed irregular long crested seas. The catamaran's sea state is: $H_s=0.3m$ and $T_p=1.48s$, where $H_s$ is the significant wave height and $T_p$ is the peak wave period. The DTBM's vessel sea state is composed of oblique waves with advancing direction at $30º$ with respect to the longitudinal direction, $H_s=10.66m$ and $T_p=13.4s$. This corresponds to a World Meteorological Organization (WMO) sea state code 8. 

\begin{table}[]
\centering
\def\arraystretch{4}\tabcolsep=5pt
\begin{tabular}{|c|c|}
\toprule
Variable & Formula \\ \hline
Wave Profile & $ \zeta (x_1,t)= \sum_{n=1}^N a_n cos[k_n(x_1-c_n t)+\varepsilon_n] $ \\ \hline
Horizontal Velocity & $u(x_1,x_2,t)=\sum\limits_{n=1}^N \cfrac{a_n \omega_n}{sinh(k_n h)} cosh[k_n(x_2+h)]cos[k_n(x_1-c_n t)+\varepsilon_n]$ \\ \hline
Vertical Velocity & $v(x_1,x_2,t)=\sum\limits_{n=1}^N \cfrac{a_n \omega_n}{sinh(k_n h)} sinh[k_n(x_2+h)]sin[k_n(x_1-c_n t)+\varepsilon_n]$ \\ \hline
Horizontal Acceleration & $\dot{u}(x_1,x_2,t)=\sum\limits_{n=1}^N \cfrac{a_n \omega_n}{cosh(k_n h)} cosh[k_n(x_2+h)]sin[k_n(x_1-c_n t)+\varepsilon_n]$ \\ \hline
Vertical Acceleration & $\dot{v}(x_1,x_2,t)=\sum\limits_{n=1}^N \cfrac{a_n \omega_n}{sinh(k_n h)} cosh[k_n(x_2+h)]cos[k_n(x_1-c_n t)+\varepsilon_n]$ \\ \hline
Dynamic Pressure & $p(x_1,x_2,t)=\sum\limits_{n=1}^N \cfrac{a_n\rho g}{sinh(k_n h)} cosh[k_n(x_2+h)]cos[k_n(x_1-c_n t)+\varepsilon_n]$ \\ \hline \hline 
\bottomrule
\end{tabular}
\medskip
\caption{Formulation for irregular long crested seas, with ($x_1,x_2$) spatial location in the mean plane of the surface elevations.}
\label{tab:irr}
\end{table}

\begin{figure}[h!] 
    \centering
        \includegraphics[width=0.95\columnwidth]{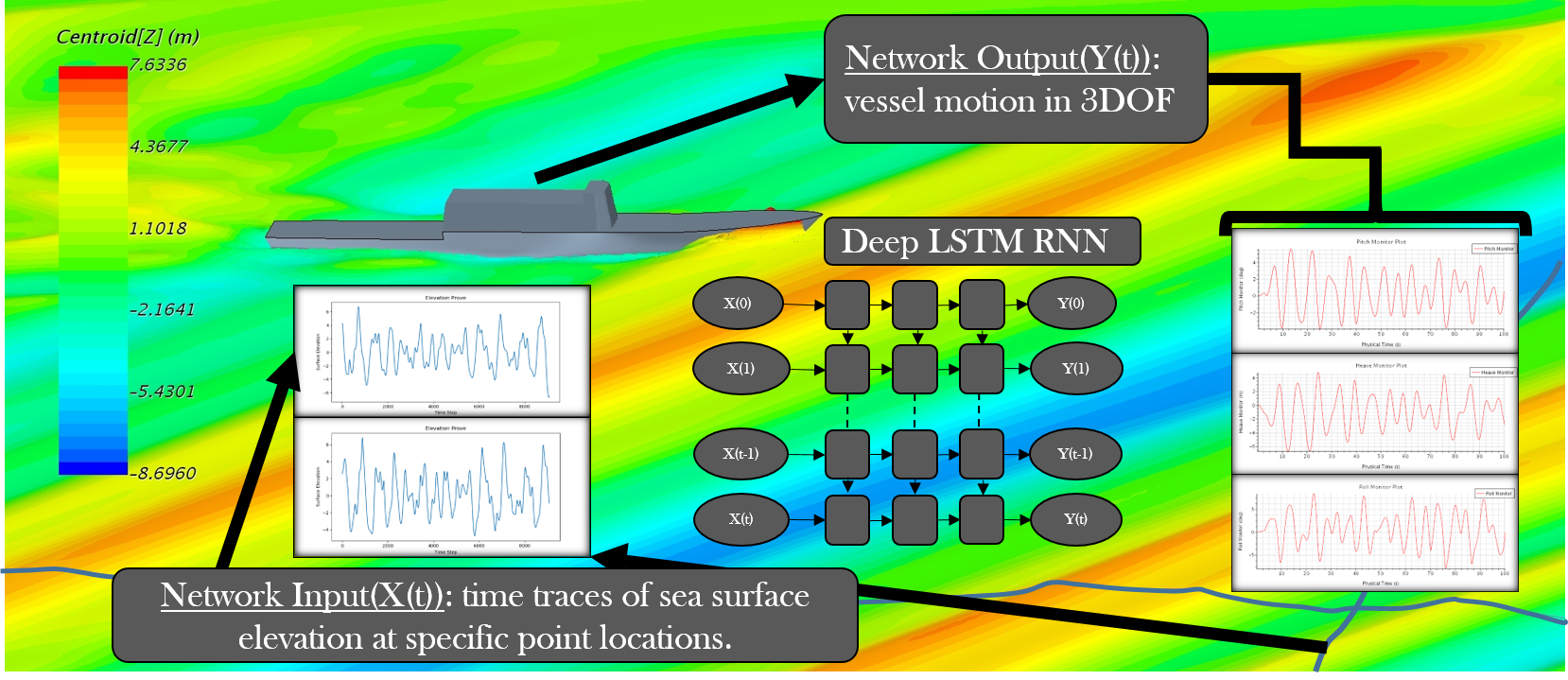}
    \medskip
    \caption{\textit{Schematic of the physical problem simulated and inputs and outputs of the deep LSTM RNN.} The inputs for training are sea surface elevations in the form of time series, while the corresponding outputs are the vessel motions. Sea surface elevations are recorded at specific point locations that can be chosen from lines over the free surface. Vessel motions in the training cases are obtained from an URANS solver. Shown here as inputs ($X(t)$) are two unseen surface elevations, which serve as test cases in our simulation example for the DTMB vessel.}
\label{fig:MasterFigure}
\end{figure}   

\section{Approximation of Continuous Functionals by Neural Networks \& Application to Dynamical Systems}
\label{sec:Sec_modeling}

Our approach is inspired by the theoretical work in~\cite{Functionals}, which proves that NNs can approximate arbitrarily well continuous functionals. A further step to nonlinear functional mapping (from a space of functions into real numbers) would be nonlinear operator mapping (from a space of functions into another space of functions). The theoretical basis to this is given in \cite{chen1995universal} and  good example of it is DeepONet \cite{lu2019deeponet}.  Some related references about these results are \cite{hornik1989multilayer,hornik1991approximation,cybenko1989approximation,sandberg1992approximation,kreinovich1991arbitrary}. LSTM type networks provide robust modeling for complex dynamic physical systems. Nevertheless, the data analyzed cannot be guaranteed to comply with all the assumptions under which the theorems presented in \cite{Functionals} are proven. Specifically, we refer to the continuity of the functions being fed to the functional that models the physical system and the compactness of the sets on which they are defined, giving reason to believe that these results can be further generalized. % {\bf XXX - what do you mean here? - Actually, In one of our discussions you mentioned that NN can also represent discontinuous functions if they can be defined piece-wise (or something similar). I was actually thinking about something similar here, fluid-structure interaction in our problem happens at different flow regimes that I believe could be defined piece-wise. Has this been proven for functionals as it has been proven for functions?}.

The results presented in \cite{Functionals} can be summarized as follows. Given very mild conditions, a functional defined on a compact set in $C[a,b]$ or $L^p[a,b]$ (spaces of infinite dimensions) can be approximated arbitrarily well by a neural network with just one hidden layer.  Particularly, given U a compact set in $C[a,b]$, $\sigma$ (a bounded sigmoidal function) and $\cal F$ a continuous functional defined on U, then $\forall u \in U, {\cal F} (u)$ can be approximated by 
\normalsize

\begin{equation}
\label{eq:functional}
{\cal F} (u) = \sum _ { i = 1 } ^ { N } c _ { i } \sigma \left( \sum _ { j = 0 } ^ { m } \xi _ { i , j } u \left( x _ { j } \right) + \theta _ { i } \right) .
\end{equation}
\normalsize
In the above expression, $c_i, \xi_{ij}, \theta_{i}$ are real numbers and $u(x_j)$ is the value that $u$ takes at $x_j$.

The stated theorem and the explicit expression provided can have a very wide impact on the foreseeable applications of neural networks to model dynamical systems. We have adopted this framework and  tested it on realistic applications in this paper for first time.

In our application examples, we view the output of the dynamical system (vessels with unsteady motions) as a functional of a forcing term (ocean waves). The results presented suggest that the joint application of dynamic physical models and RNN already lead to large computational savings when the physical models are complex to simulate. For the cases simulated in this paper, given the desired accuracy, we have seen that we can save huge computational simulation time, spent on producing the dataset that is designed to characterize the motions of the vessel. Furthermore, the obtained surrogate model can potentially be used to create a digital twin of the real vessel to ensure operability and safety in extreme weather conditions.%{\bf xxx - what do you mean? - The model can potentially be a digital twin of the vessel and it can be used as such. Should I rephrase?}

First, we model the two degree-of-freedom (2DOF)  (heave \& pitch) motions of a catamaran vessel  for nonlinear 5th-order regular waves (some results are presented in \cref{fig:LSTM}). Subsequently, we consider irregular stochastic waves that represent real life sea-states (some results presented in \cref{fig:Irreg,fig:ConvIrreg,fig:ConvIrreg2,fig:ConvIrregLong}). Lastly, we consider the three degrees-of-freedom (heave, pitch \& roll) motion approximation of a notional DTMB battleship  (\cref{fig:13}). In doing this, we believe that we have achieved a state-of-the-art generalization of motions that are largely governed by the Navier-Stokes equations. A brief discussion of the physics is presented in \cref{sec:cfd}, and more details can be found in \cite{del2018influence}.

\subsection{Representation of Dynamical Systems with Functionals}

We state the extension of \cref{eq:functional} to dynamical systems (taken from \cite{Functionals}) that presents the key aspects which guarantee the learnability of a dynamical system.
The \textit{first} assumption is that the dynamical system can be modeled with a continuous functional, defined in a compact set. %This is already is difficult to prove in both of the vessels analyzed. The results for both vessels are very satisfactory. 
The \textit{second} assumption is that the map that is used to represent functionals for the dynamical system can be performed with a windowing operator. Stating this in a formal way, we assume that $X_1$ and $X_2$ are sets in $\mathbf { R } ^ { q _ { 1 } }$ and $\mathbf { R } ^ { q _ { 2 } }$ -valued functions defined in $\mathbf{R}^n$. The dynamical system we study is viewed as map from $X_1$ to $X_2$, such that $\forall$ $u \in X_1$, $Gu = v \in X_2$. We can define an n-dimensional windowing operator $W$ for $x \in X$ centered at $\alpha$ and with width $2a$:

\[
\left( W _ { \alpha , a } x \right) ( \beta ) = \left\{ \begin{array} { l l } { x ( \beta ) } & { \text { if } \beta \in \Gamma _ { \alpha , a } } \\ { 0 } & { \text { if } \beta \notin \Gamma _ { \alpha , a } } \end{array} \right.
.\]
Using this window operator, we can restrict a non-empty set $U$ of $X_1$, $U _ { \alpha , a } = \left\{ \left. u \right| _ { \Gamma _ { \alpha , a } } , u \in U \right\}$. This reads as, $\left. u \right| _ { \Gamma _ { \alpha , a } }$ is the restriction of $u$ to $\Gamma _ { \alpha , a }$.

We consider that a map $G$ from $X_1$ to $X_2$ has \textit{approximately finite memory} if $\forall \epsilon > 0$ there is $a>0$ such that:

\normalsize
\[
\left| ( G u ) _ { j } ( \alpha ) - \left( G W _ { \alpha , a } u \right) _ { j } ( \alpha ) \right| < \epsilon , \quad j = 1 , \ldots , q _ { 2 } \quad \forall \alpha \in \mathbf{R}^n, u \in U
.\]
\normalsize

The above statement has a very deep impact on the learnability of the dynamical process. It limits the extension of the proof of approximation of functional to those dynamical systems for which we do not need to  know all the history of states in order to approximate them. Furthermore, to give an explicit expression of the functional, the following assumptions are necessary:

\begin{enumerate}
\item
If $u \in U$, then $\left. u \right| _ { \Gamma _ { \alpha , a } } \in U$, $\forall\alpha \in \mathbf { R } ^ { n } , a > 0$.
\item
$\forall$ $\alpha \in \mathbf { R } ^ { n } , a > 0 , U _ { \alpha , a }$ is a compact set in $C _ { V } \left( \prod _ { k = 1 } ^ { n } \left[ \alpha _ { k } - a _ { k } , \alpha _ { k } + a _ { k } \right] \right)$ or a compact set in $L _ { V } ^ { p } \left( \prod _ { k = 1 } ^ { n } \left[ \alpha _ { k } - a _ { k } , \alpha _ { k } + a _ { k } \right] \right)$, where $V$ stands for $\mathbf{R}^{q_1}$.
\item
Then, if we let $( G u ) ( \alpha ) = \left( ( G u ) _ { 1 } ( \alpha ) , \ldots , ( G u ) _ { q _ { 2 } } ( \alpha ) \right)$, consequently each $( G u ) _ { j } ( \alpha )$ will be a continuous functional defined over $U _ { \alpha , a }$, with the corresponding topology in $C _ { V } \left( \prod _ { k = 1 } ^ { n } \left[ \alpha _ { k } -\right. \right. a _ { k } , { \alpha } _ { k } + a _ { k } ]$  or $L _ { V } ^ { p } \left( \prod _ { k = 1 } ^ { n } \left[ \alpha _ { k } - a _ { k } , \alpha _ { k } + a _ { k } \right] \right)$.
\end{enumerate} 

Given the above results and following the process given in reference \cite{Functionals}, we can find the extension of \cref{eq:functional} to dynamical systems in the following theorem:

\textbf{Theorem}: If $U$ and $G$ satisfy all the assumptions (1-3) made previously, and $G$ is of \textit{approximately finite memory}, then $\forall \epsilon > 0$, $\exists a > 0$, $m$ a positive integer, $(m+1)^n$ points in $\prod _ { k = 1 } ^ { n } \left[ \alpha _ { k } - a _ { k } , \alpha _ { k } + a _ { k } \right]$, $N$ a positive integer, constants $c_i(G,\alpha, a)$ that only depend on $G,\alpha, a$, and $q _ { 2 } \times ( m + 1 ) ^ { n }$ - vectors $\xi _ { i }$, $i = 1 , \dots , N$, such that:
\normalsize
\[ 
\left| ( G u ) _ { j } ( \alpha ) - \sum _ { i = 1 } ^ { N } c _ { i } ( G , \alpha , a ) \sigma \left( \overline { \xi } _ { i } \cdot \overline { u } _ { q _ { 1 } , n , m } + \theta _ { i } \right) \right| < \epsilon , \quad j = 1,2 , \ldots , q _ { 2 }
.\]
\normalsize
To conclude, we would like to place emphasis on the assumption of \textit{approximately finite memory}. This assumption provides the blocks to build the functional approximation as a sum of functionals defined in the subsets given by the window operator, previously defined. During the empirical analysis, we benchmark against a case for which we hope that \textit{approximately finite memory} will allow representing the functional arbitrarily well from the subsets given by the window operator.
%%%%%%%%%%%%%%%%%%%%%%%%%%%%%%%%%%%%%%%%%%%%%%%%%%%%%%%%%%%%%%%%%%%%%%%%%%%%%%%%%%%%%%%%%%%%%%%%%%%
\section{A Brief Overview of the Machine Learning Algorithm}

RNNs are a generalization of feedforward neural networks that provide a better architecture for storing key aspects of sequences. Furthermore, they are able to maintain a notion of \textit{state}. This \textit{state} evolves as a larger fragment of the sequence is 'seen' by the network. Furthermore, we compute and/or predict sequences considering this \textit{state} with some information that becomes available to the network. Ongoing research has proven RNNs to be powerful but difficult to train. The main difference between feed-forward neural networks and RNNs is that RNNs use the same functional mapping to transform the previous state, providing parameter sharing. Since basic RNNs suffer from vanishing gradient problems, and given that the time series used in this paper are quite long, we have to use gated units, i.e., LSTM and GRU cells. The gates in these units are used to control the flow of information that passes through the state of the cell. For the first part, after a convergence study of the properties of the networks (\cref{fig:LSTM}), we have found LSTMs to be more expressive. We have also tested other feed-forward architectures such as NARX (Nonlinear Auto-Regressive with Exogenous input); however, they have been discarded because they are unable to 'forget', which results in an ever increasing residual error that propagates and compromises the performance of the NN when predicting long sequences.

The chosen type of gated cell (LSTM) was introduced two decades ago \cite{LSTM} and has now gained popularity in the context of language modeling. However, we believe its full potential regarding time series modeling is relatively underappreciated at the moment.
The formulation of LSTM cells is as follows \cite{LSTM,HandOnML}. Let $c_t$ the internal memory and $h_t$ the visible state. Consequently, the state that evolves is the pair $[c_t,h_t]$. The forget gate $f_t$, as its name indicates, controls what to forget from the memory cell. The input gate $i_t$, controls what to read out of the memory cell into the visible state $h_t$, via the output gate $o_t$. Finally, the memory cell $c_t$ evolves by the addition of inputs that come from the forget and input gates.
\normalsize
\[f _ { t } = \sigma \left( W ^ { f , h } h _ { t - 1 } + W ^ { f , x } x _ { t } \right) \quad \text { (forget gate)}\]
\[i _ { t } = \sigma \left( W ^ { i , h } h _ { t - 1 } + W ^ { i , x } x _ { t } \right) \quad \text { (input gate) }\]
\[o _ { t } = \sigma\left( W ^ { o , h } h _ { t - 1 } + W ^ { o , x } x _ { t } \right) \quad \text { (output gate) }\]
\[c _ { t } = f _ { t } \odot c _ { t - 1 } + i _ { t } \odot \sigma_h \left( W ^ { c , h } h _ { t - 1 } + W ^ { c , x } x _ { t } \right) \quad \text {(memory cell)}\]
\[h _ { t } = o _ { t } \odot \sigma_h \left( c _ { t } \right) \quad  \text {(visible state)}\]
\[p _ { t } = \operatorname { softmax } \left( W ^ { o } h _ { t } \right)
.\]
\normalsize

The objective function to minimize during the training is the mean-squared-error (MSE):

\begin{equation}
    \mathrm{MSE}=\frac{1}{n} \sum_{i=1}^{n}\left(Y_{i}-\hat{Y}_{i}\right)^{2}
    ,
\end{equation}

where $Y$ is the vector of observed values and $\hat{Y}$ are the predicted values.

\section{Nonlinear Functional Approximation for Modeling Nonlinear Motions}

To construct the nonlinear functional approximation for predicting the motion dynamics, we have designed LSTM networks that approximate 2DOF or 3DOF motions of vessels advancing in irregular waves in head or oblique seas. We first present results for the catamaran vessel in mild WMO sea state 1, and subsequently we present results for the DTBM vessel in extreme WMO sea state 8.

\subsection{Catamaran Vessel}
\label{sec:Catmaran}
\begin{figure*}[h!] 
 \centering
    \begin{subfigure}[t]{0.45\textwidth}
        \centering
        \includegraphics[width=\columnwidth]{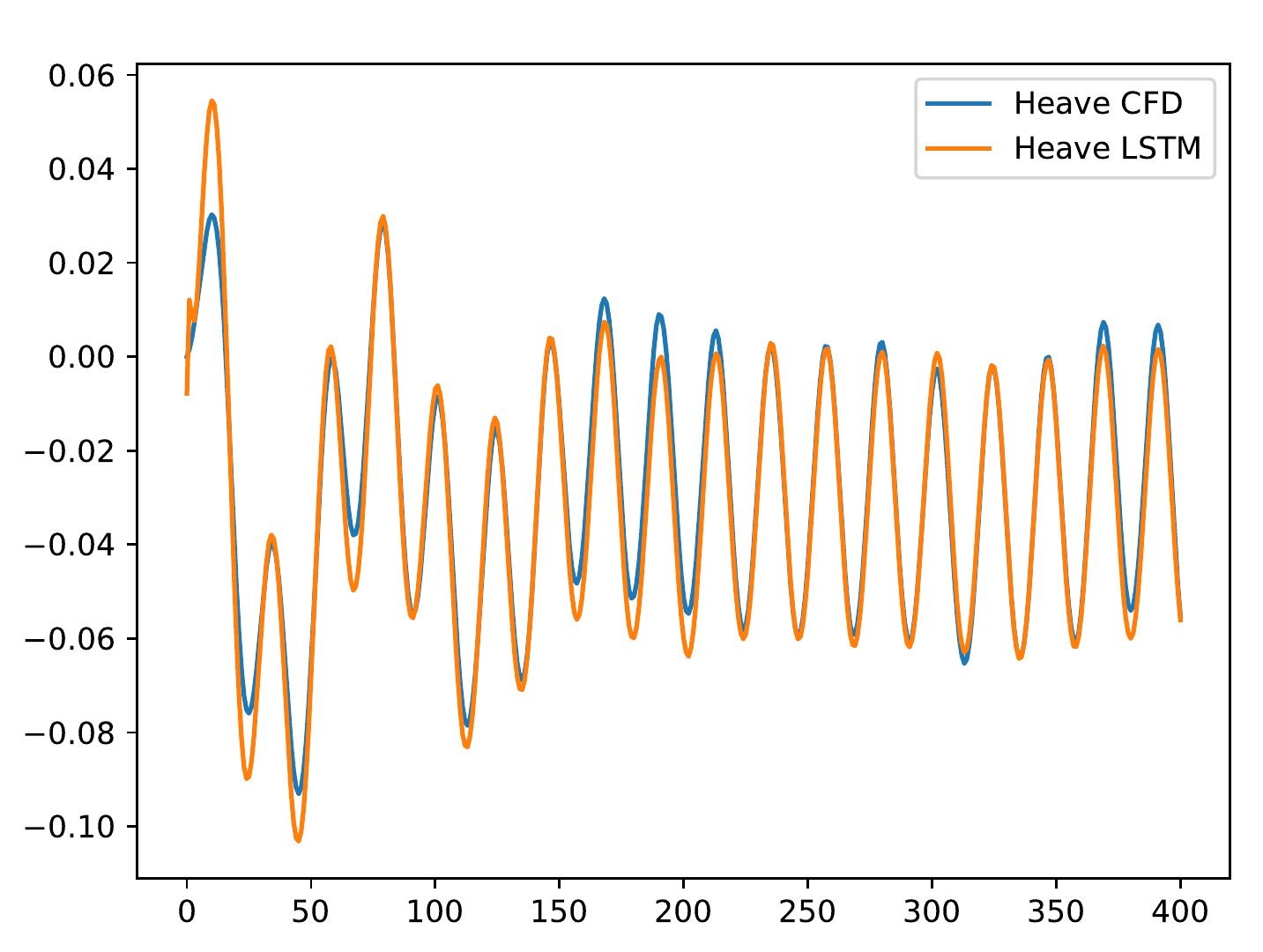}
        \caption{\scriptsize LSTM test: 20 neurons, 1 hidden layer.} 
		\label{fig:LSTMc}    
    \end{subfigure}
    ~
    \begin{subfigure}[t]{0.45\textwidth}
        \centering
        \includegraphics[width=\columnwidth]{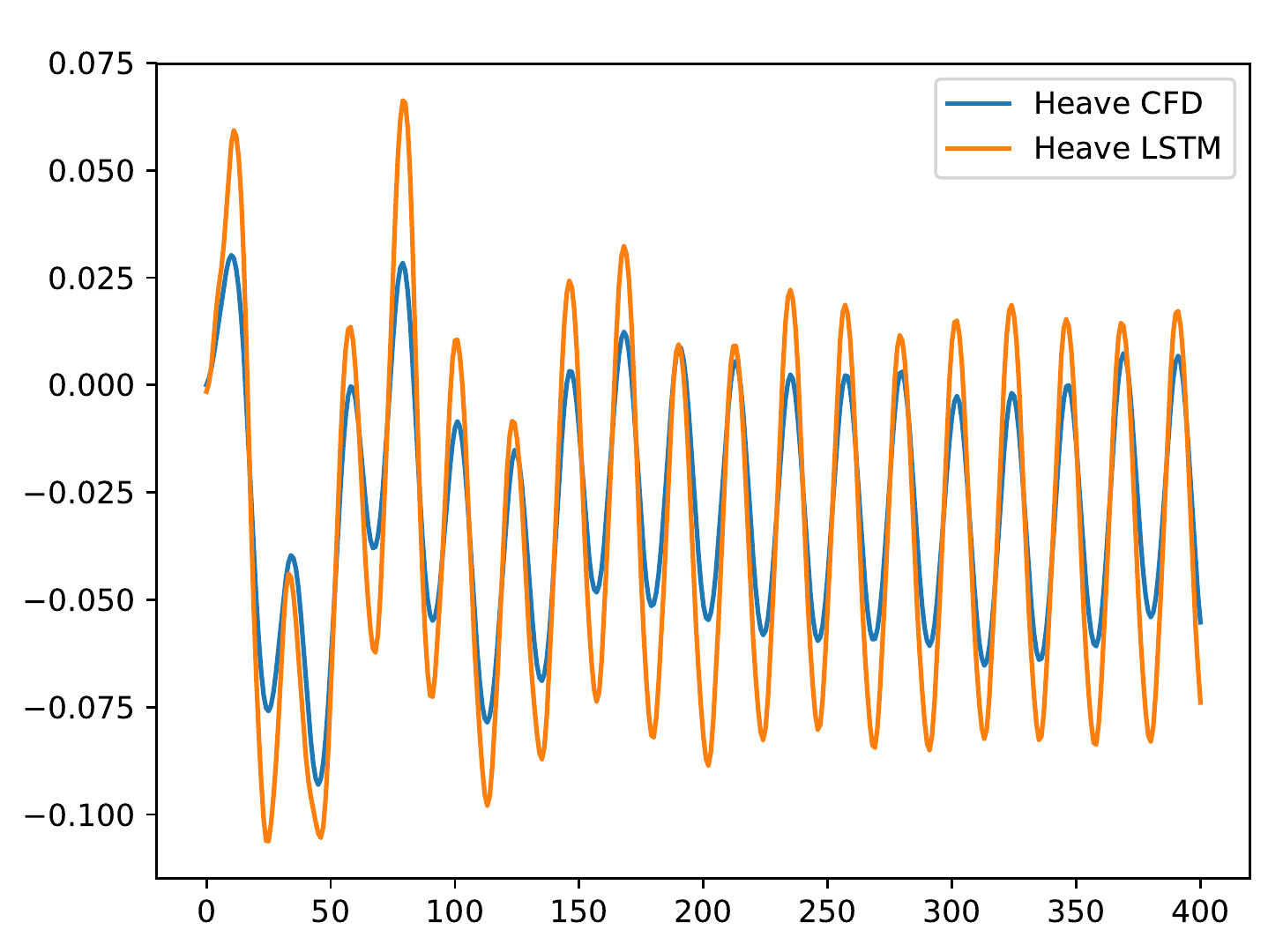}
        \caption{\scriptsize LSTM test: 5 neurons, 1 hidden layer.}
        \label{fig:LSTMd}
    \end{subfigure}%
    
    \begin{subfigure}[t]{0.45\textwidth}
        \centering
        \includegraphics[width=\columnwidth]{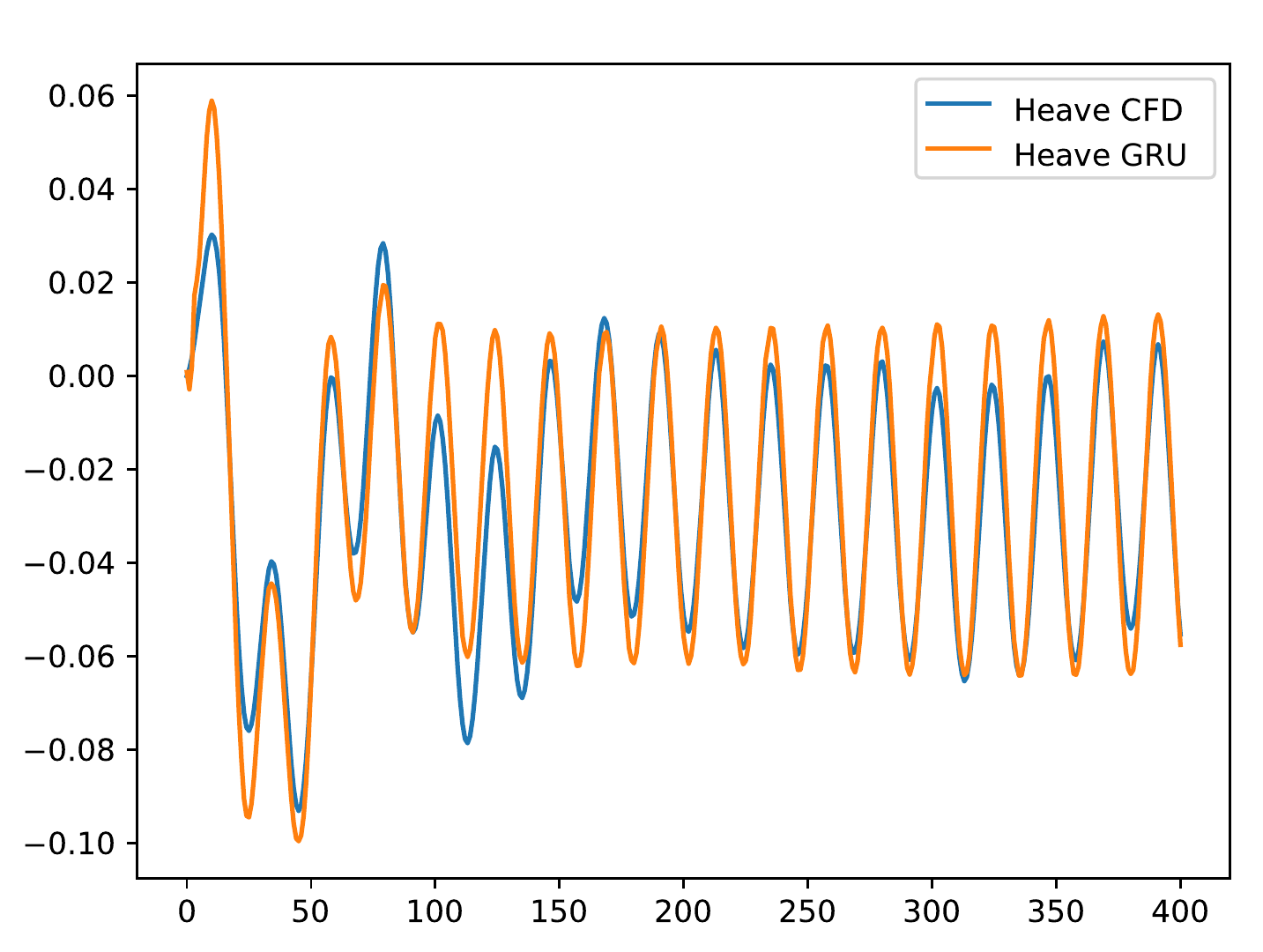}
        \caption{\scriptsize GRU test: 20  neurons, 1 hidden layer.} 
		\label{fig:GRUg}    
    \end{subfigure}
    ~
    \begin{subfigure}[t]{0.45\textwidth}
        \centering
        \includegraphics[width=\columnwidth]{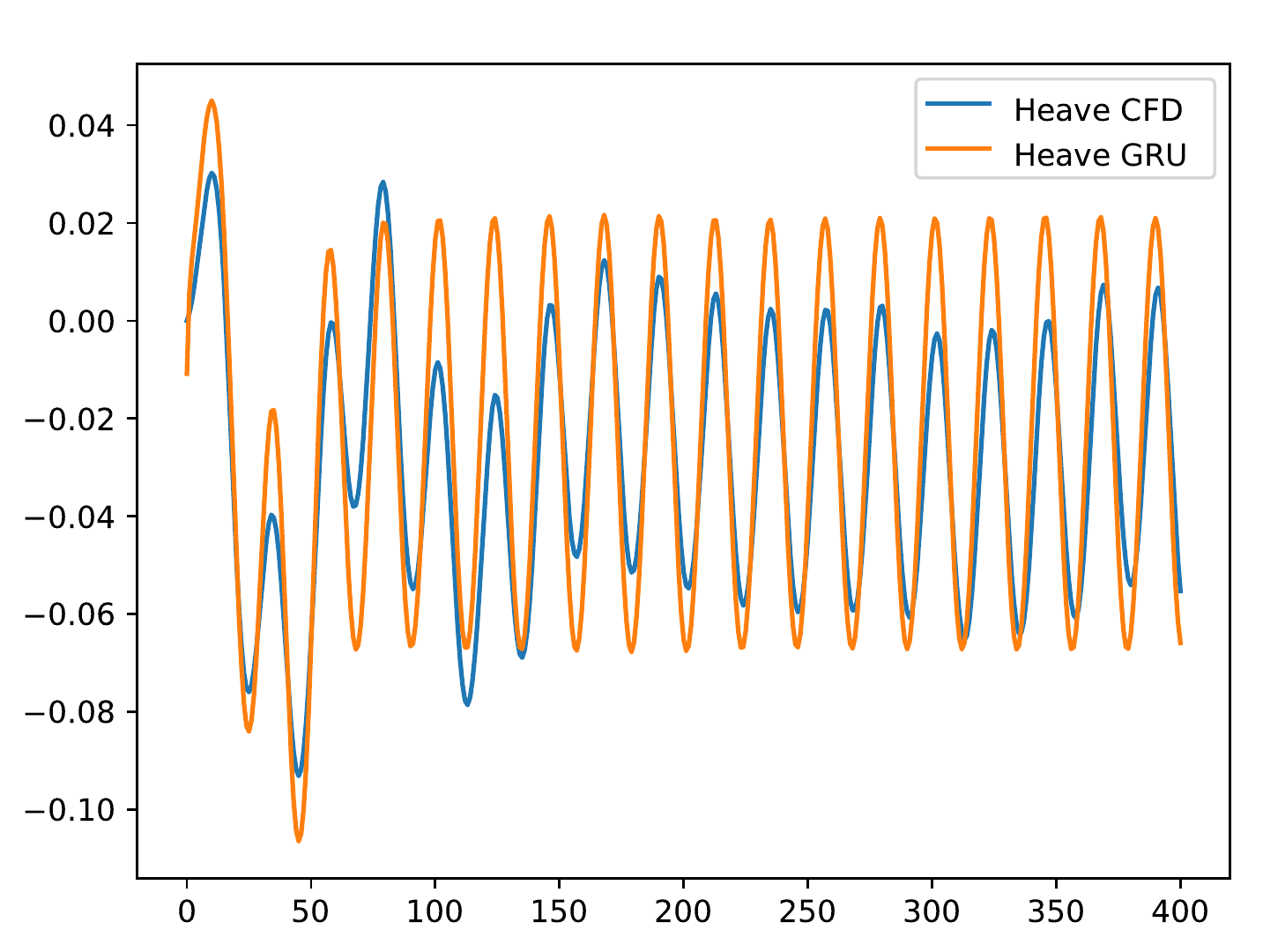}
        \caption{\scriptsize GRU test: 5 neurons, 1 hidden layer.}
        \label{fig:GRUh}
    \end{subfigure}%
    
    \medskip
    \caption{\textit{Comparison of LSTM and GRU for a 5th-order Stokes regular wave (wave amplitude is $0.15m$) for the catamaran vessel.} The data-set is composed of 5 waves of varying amplitude. The first 4 waves are used as training cases (for 20000 steps) while the last wave (shown here) is used for testing. Each time step corresponds to $\Delta t=0.0625s$.}
\label{fig:LSTM}
\end{figure*}

 We start with simple Stokes 5th-order waves (see results in \cref{fig:LSTM}). First, we develop benchmark cases to choose between the two most used gated RNN cells. Specifically, we have compared two different regular waves and two different network sizes (5 and 20 neurons) for one hidden layer for GRU and LSTM networks. In \cref{fig:LSTM}, we present representative results that demonstrate that LSTM networks are more expressive for our particular application, as we increase the number of neurons from 5 to 20. In the following, we adopt a LSTM network with one hidden layer for regular waves, however, we will employ deeper networks for irregular waves. 

Having chosen the type of neural network, we have tested the model capabilities with stochastic waves modeling typical sea states that the catamaran vessel may encounter. We have found that the pitch angular motion is modeled better than the vertical motion. The next step that we take is to optimize the architecture of the network augmenting its expressiveness without inducing data over-fitting.

\begin{figure*}[] 
 \centering
    \begin{subfigure}[t]{0.55\textwidth}
        \centering
        \includegraphics[width=\columnwidth]{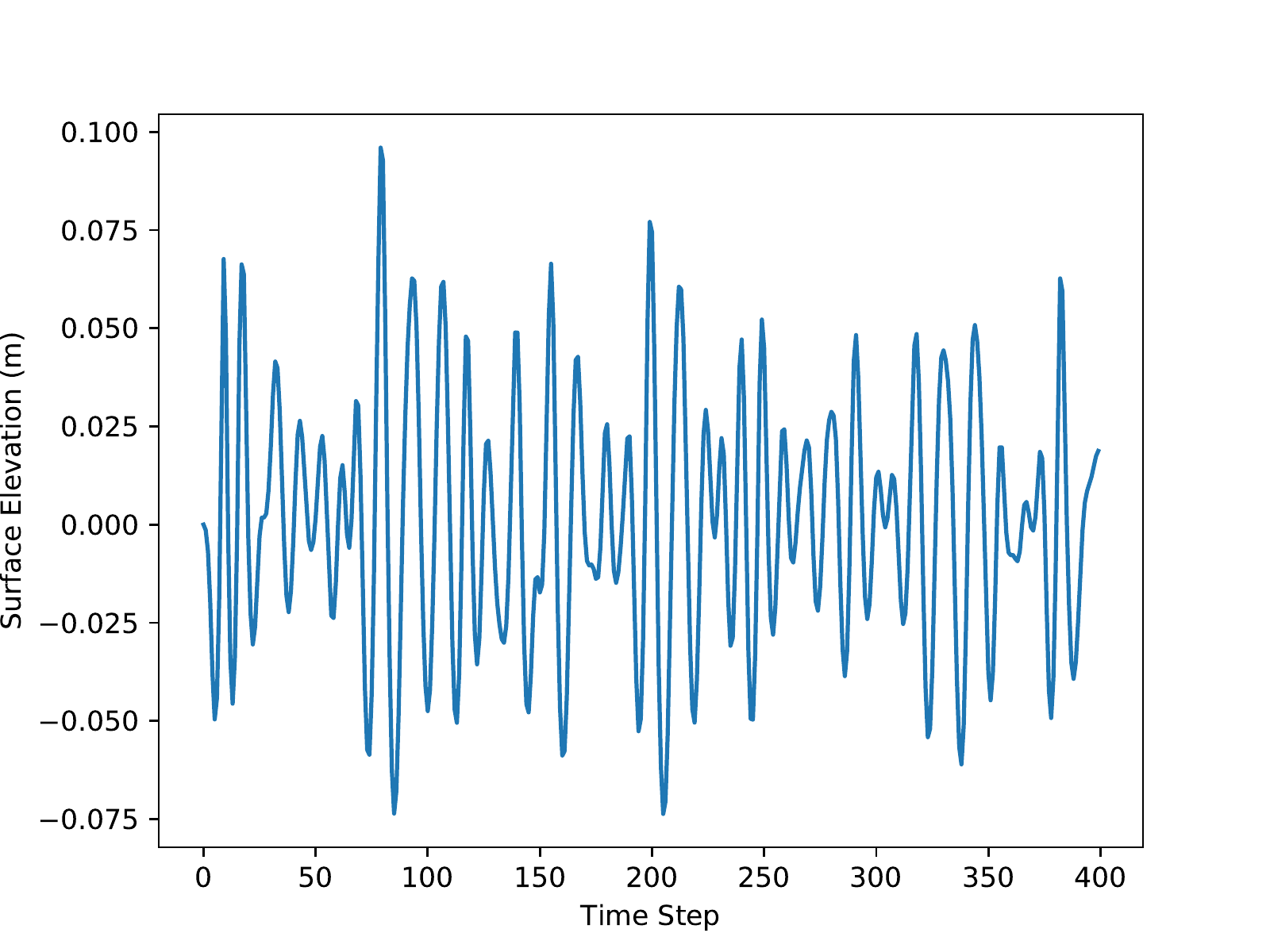}
        
        \caption{\scriptsize Network input \cref{fig:Heavea,fig:Pitchc}.} 
		\label{fig:Inputa}    
    \end{subfigure}
    
    \begin{subfigure}[t]{0.55\textwidth}
        \centering
        \includegraphics[width=\columnwidth]{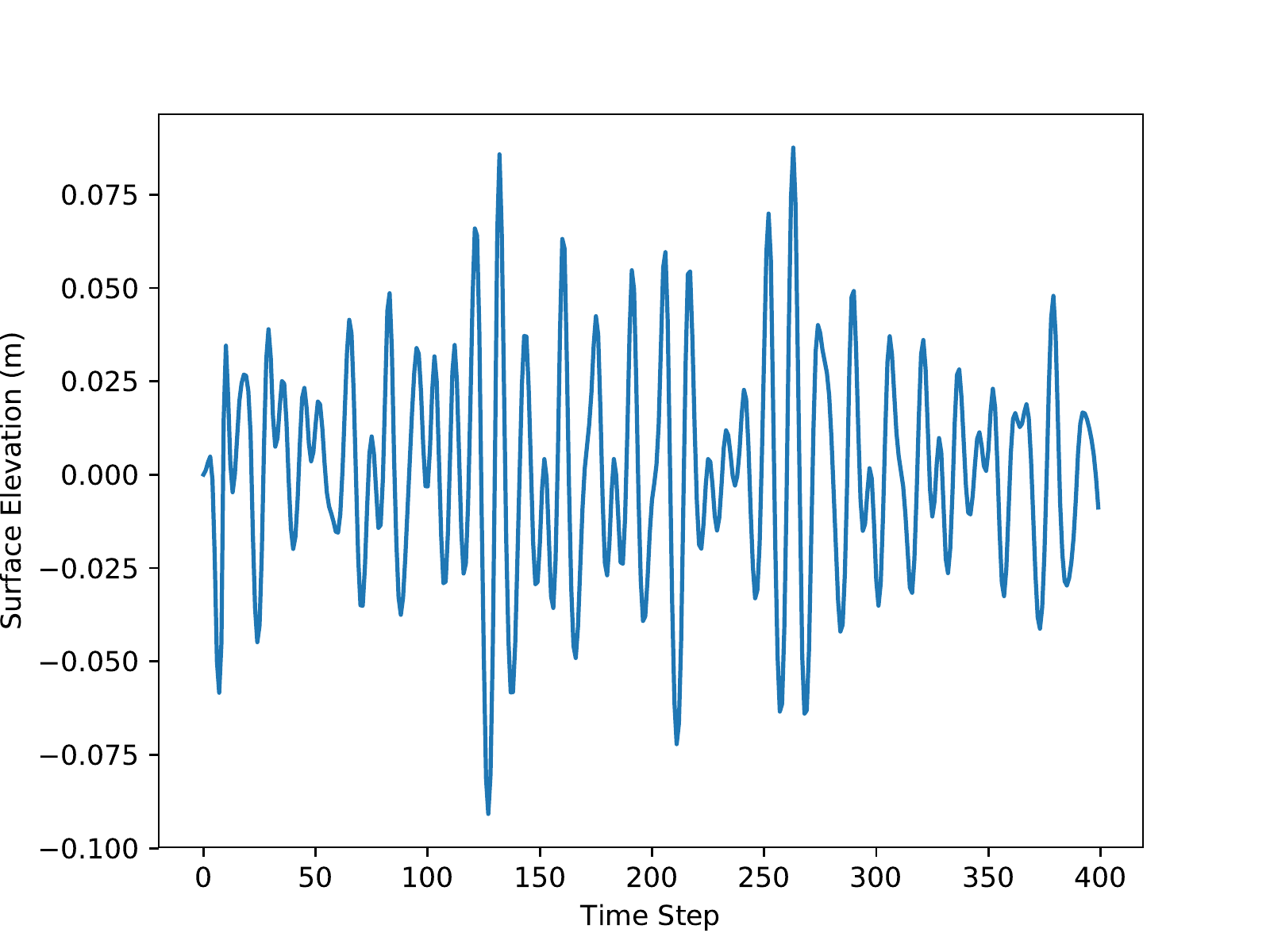}
        \caption{\scriptsize Network input \cref{fig:Heaveb,fig:Pitchd}.}
        \label{fig:Inputb}
    \end{subfigure}%
    
    \begin{subfigure}[t]{0.55\textwidth}
        \centering
        \includegraphics[width=\columnwidth]{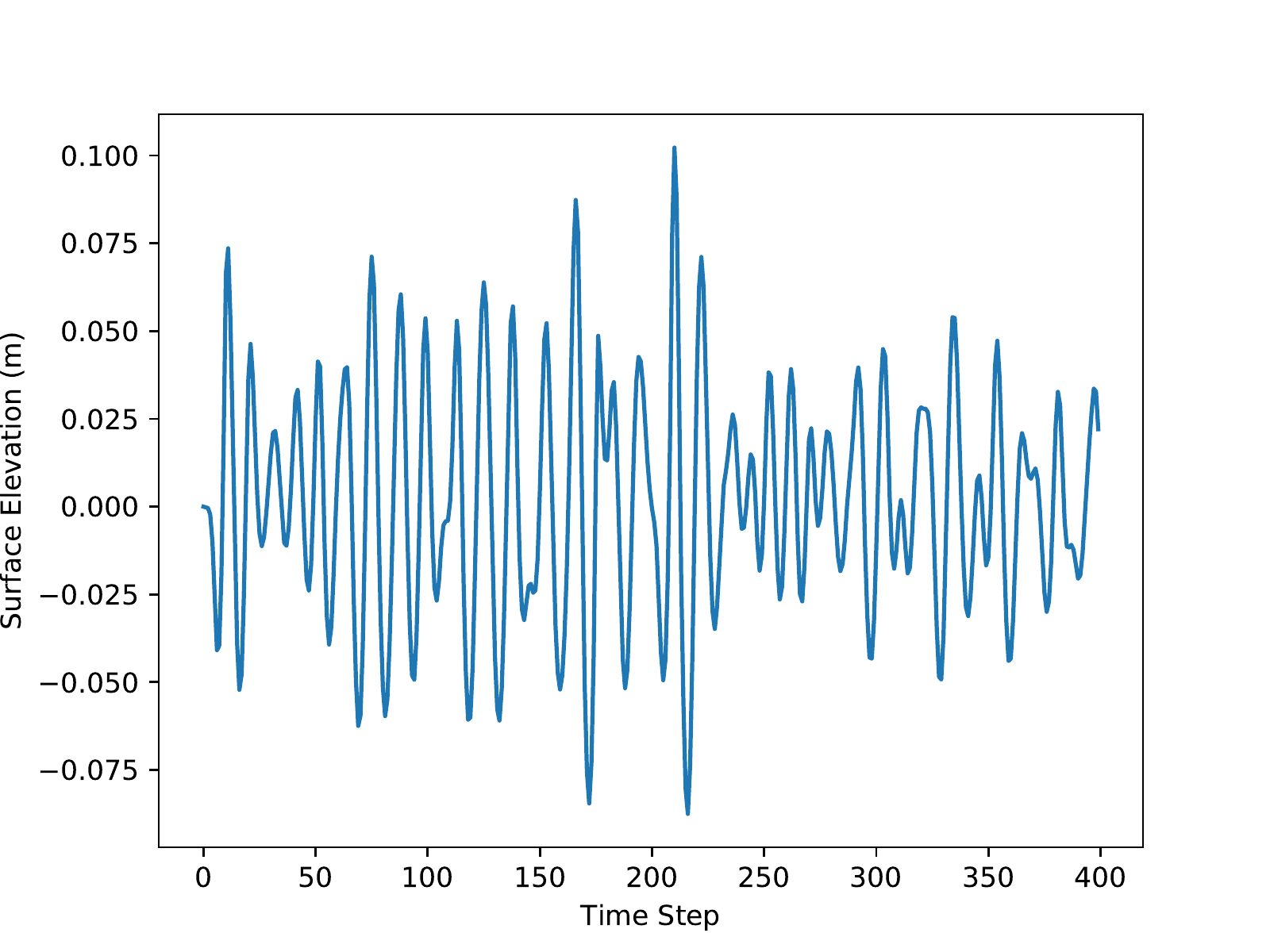}
        \caption{\scriptsize Network input \cref{fig:ConvIrreg}.}
        \label{fig:Inputc}
    \end{subfigure}%
    \medskip
    \caption{\textit{LSTM network inputs for 2DOF catamaran vessel in irregular head seas}. Each time step corresponds to $\Delta t=0.0625s$.}
\label{fig:InputIrreg}
\end{figure*}

\begin{figure*}[] 
 \centering
    \begin{subfigure}[t]{0.4\textwidth}
        \centering
        \includegraphics[width=\columnwidth]{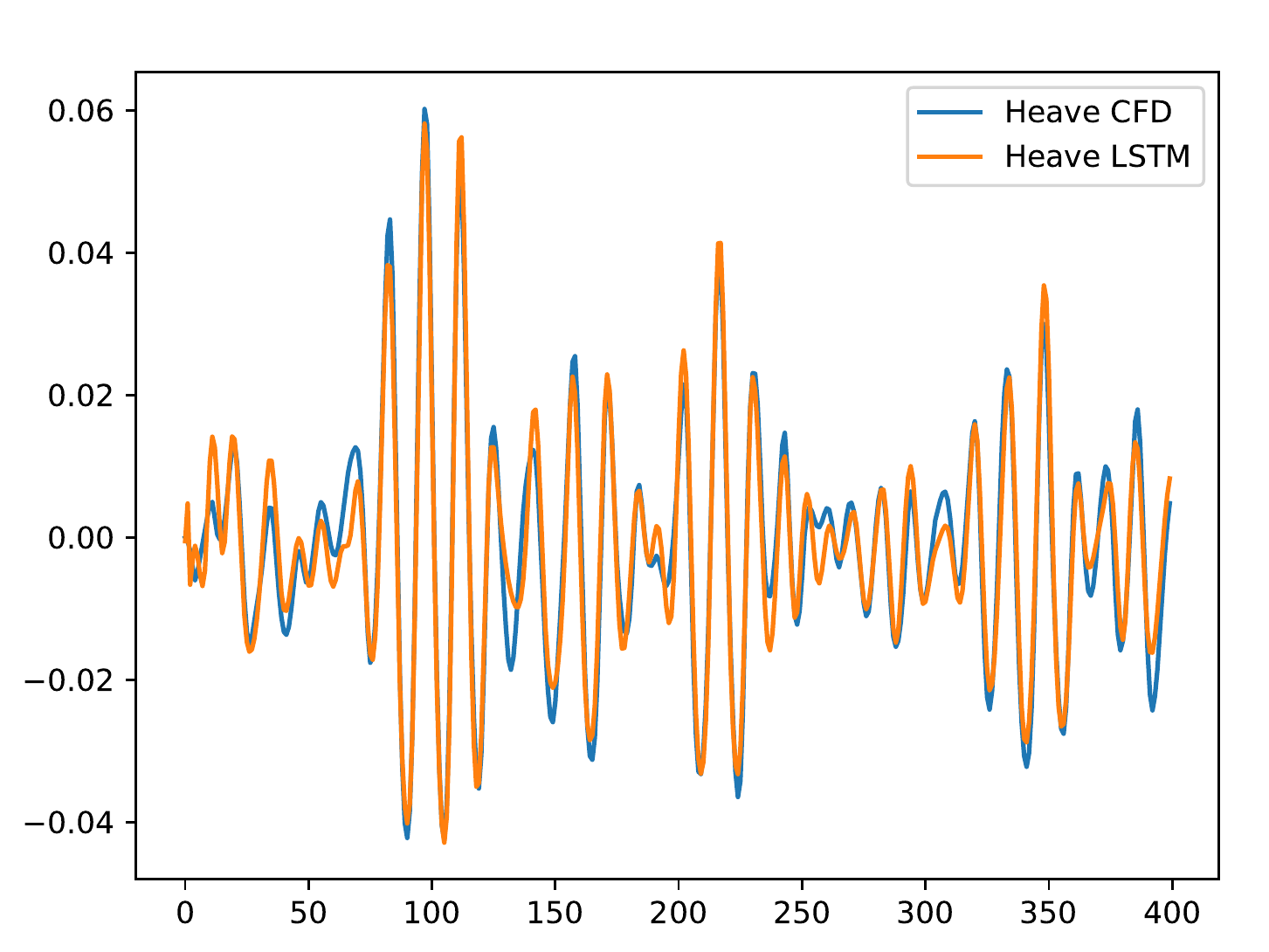}
        
        \caption{\scriptsize Vertical vessel motion. Training based on three different sequences.} 
		\label{fig:Heavea}    
    \end{subfigure}
    ~
    \begin{subfigure}[t]{0.4\textwidth}
        \centering
        \includegraphics[width=\columnwidth]{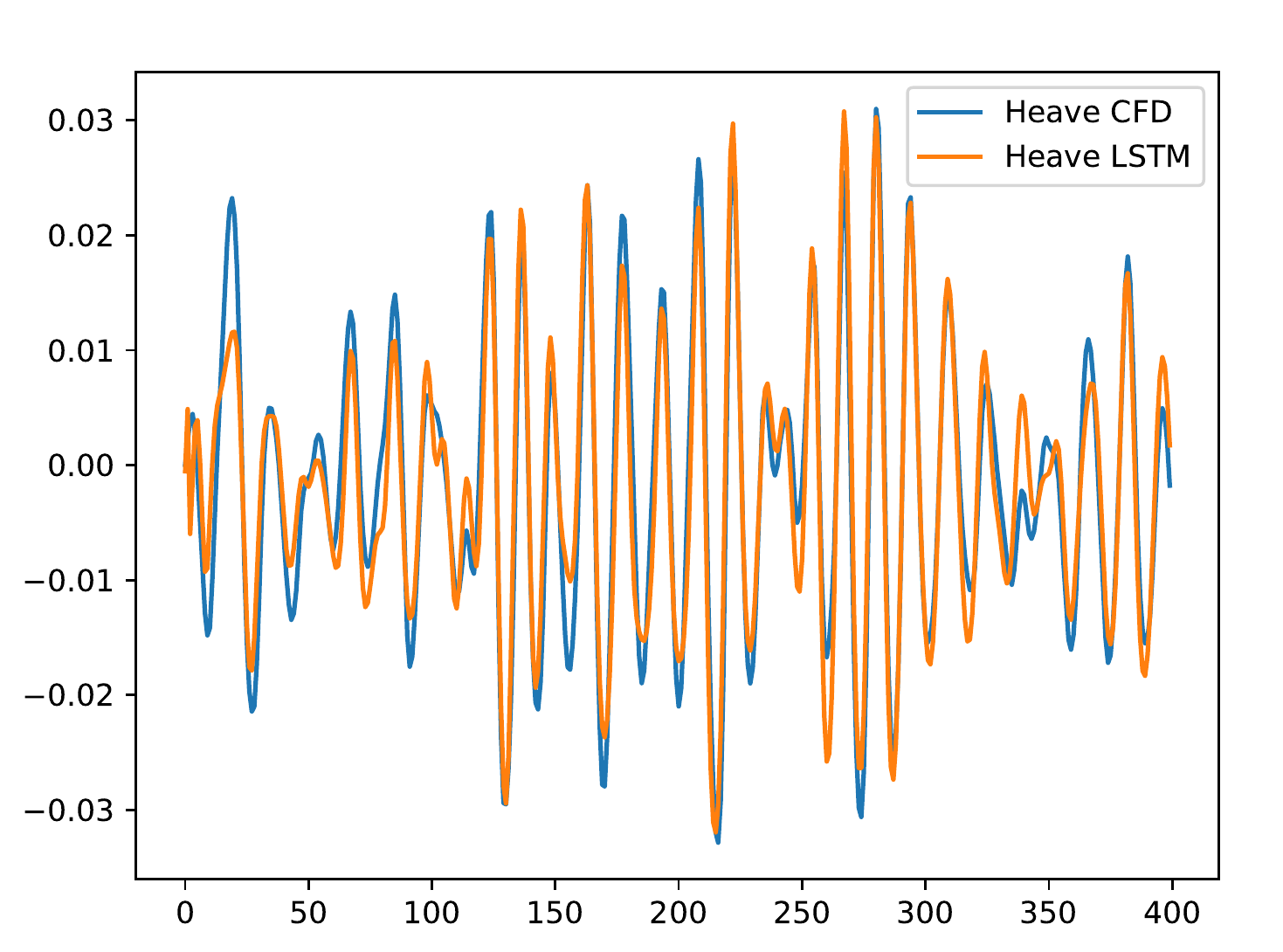}
        \caption{\scriptsize Vertical vessel motion. Testing based on the training of (a).}
        \label{fig:Heaveb}
    \end{subfigure}%
    
    \begin{subfigure}[t]{0.4\textwidth}
        \centering
        \includegraphics[width=\columnwidth]{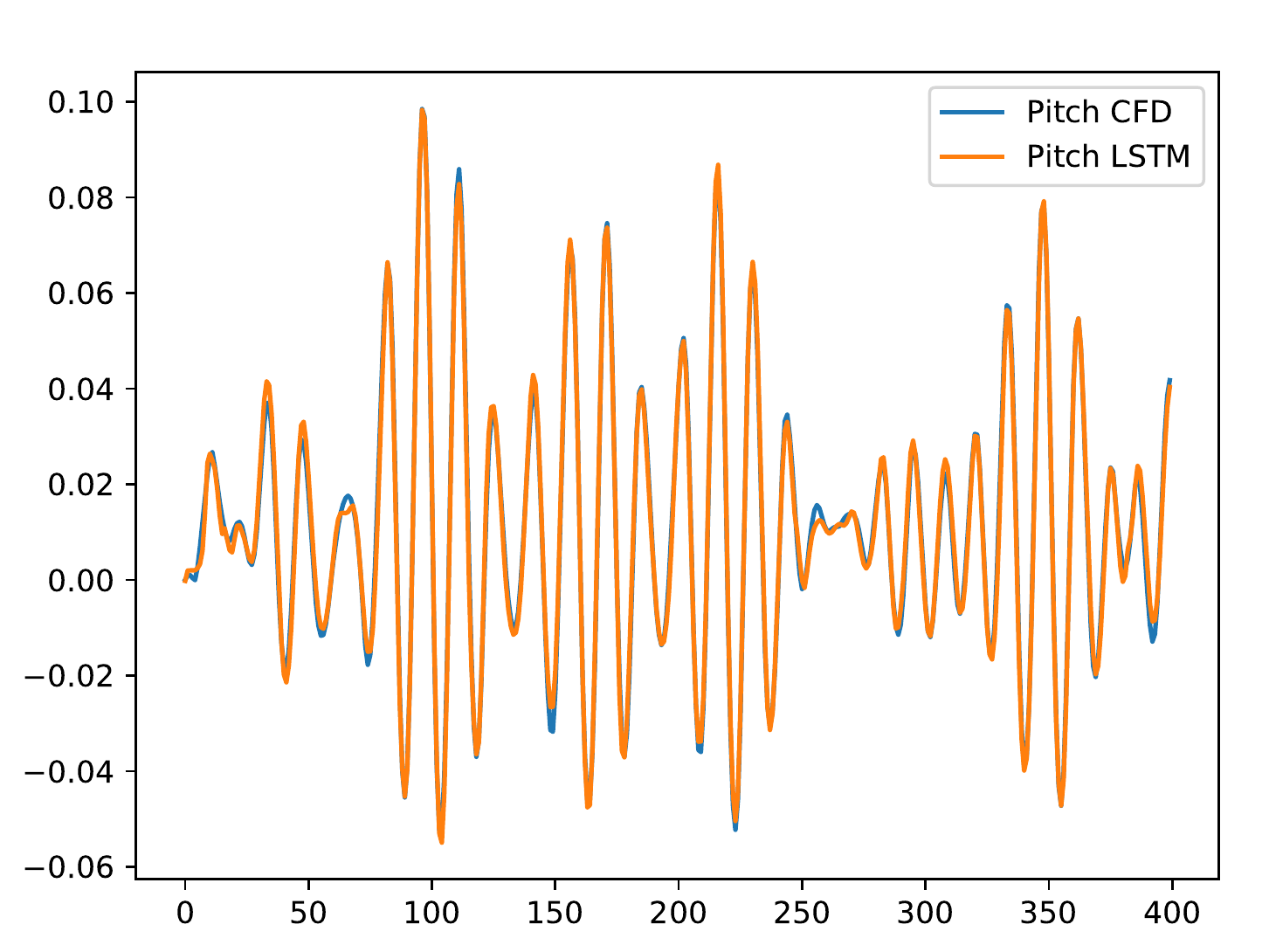}
        \caption{\scriptsize Angular vessel motion. Training based on three different sequences.} 
		\label{fig:Pitchc}    
    \end{subfigure}
    ~
    \begin{subfigure}[t]{0.4\textwidth}
        \centering
        \includegraphics[width=\columnwidth]{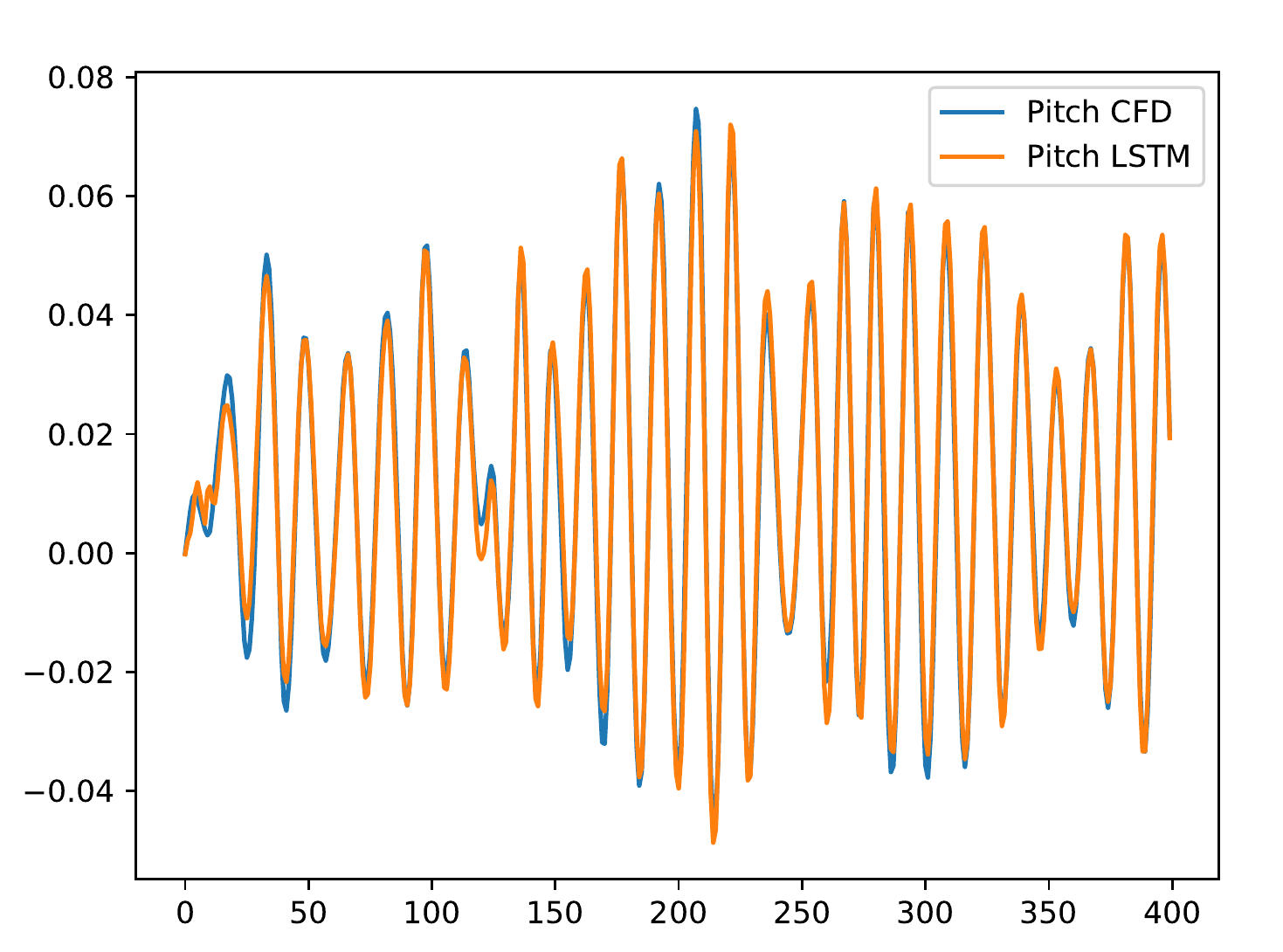}
        \caption{\scriptsize Angular vessel motion. Testing based on the training of (c).}
        \label{fig:Pitchd}
    \end{subfigure}%

    \medskip
    \caption{\textit{LSTM network (20 neurons, 1 layer) for the catamaran vessel subject to irregular waves}. The left column (a, c) shows the vertical and angular vessel motions after training with three different sea state realizations; one such realization is shown in \cref{fig:Heavea}. The right column (b, d) shows the vertical and angular vessel motions for testing given the inputs indicated in \cref{fig:Heaveb}. Each time step corresponds to $\Delta t=0.0625s$.}
\label{fig:Irreg}
\end{figure*}

\begin{figure*}[] 
 \centering
 %%%%%%%%%%%%%%%%%%%%%%%%%%%%%%%%%%%%%%%%%%%%%%%%%%%%%%%%%%%%%%%%%%%%%%%%%%%%%%%%%%%%%%%%%%%%%%%%%
    \begin{subfigure}[t]{0.32\textwidth}
        \centering
        \includegraphics[width=\columnwidth]{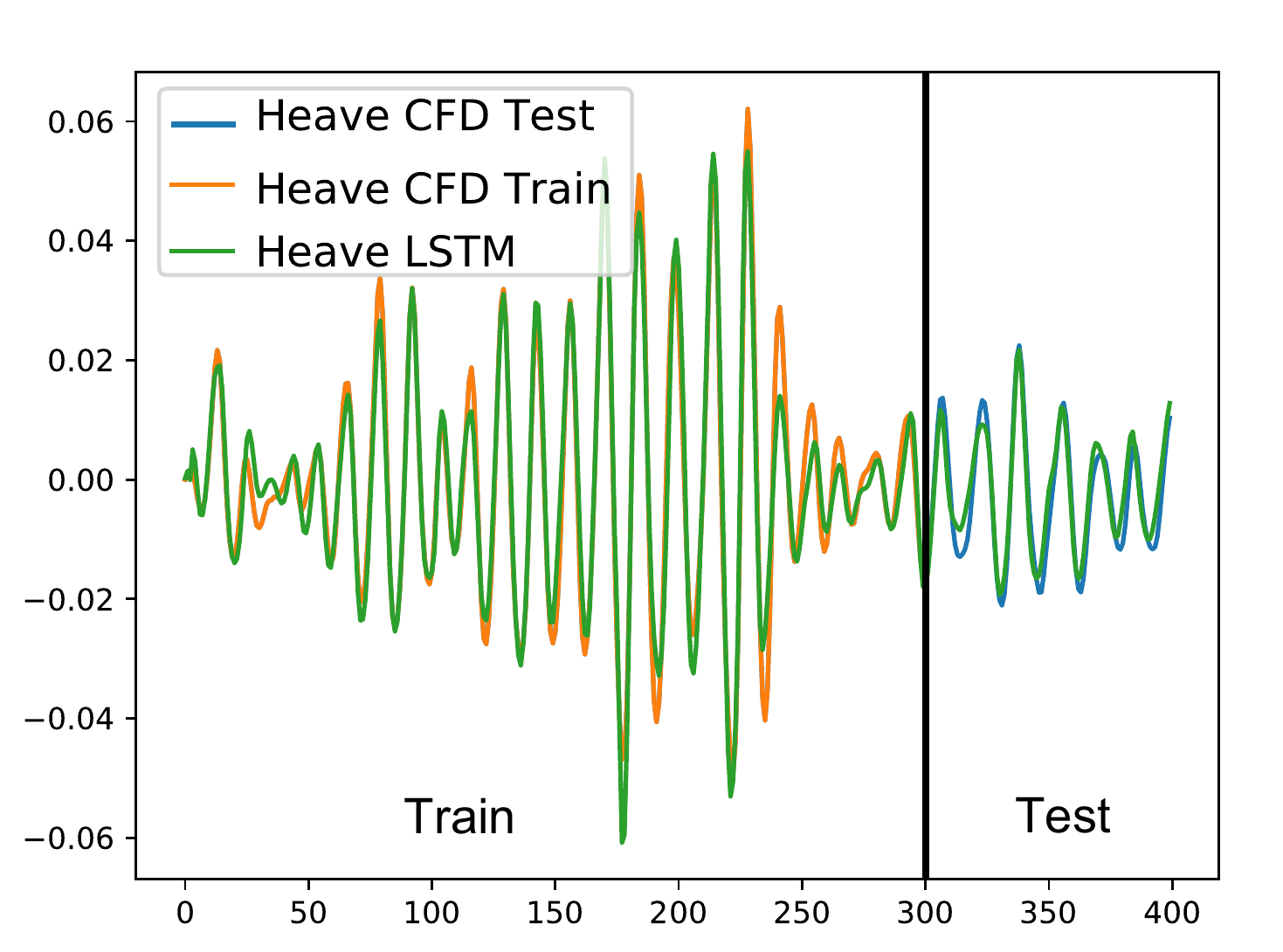}
        \caption{\scriptsize RSE= 1.7020e-05.} 
		\label{ConvIrrFig:a}    
    \end{subfigure}
    ~
    \begin{subfigure}[t]{0.32\textwidth}
        \centering
        \includegraphics[width=\columnwidth]{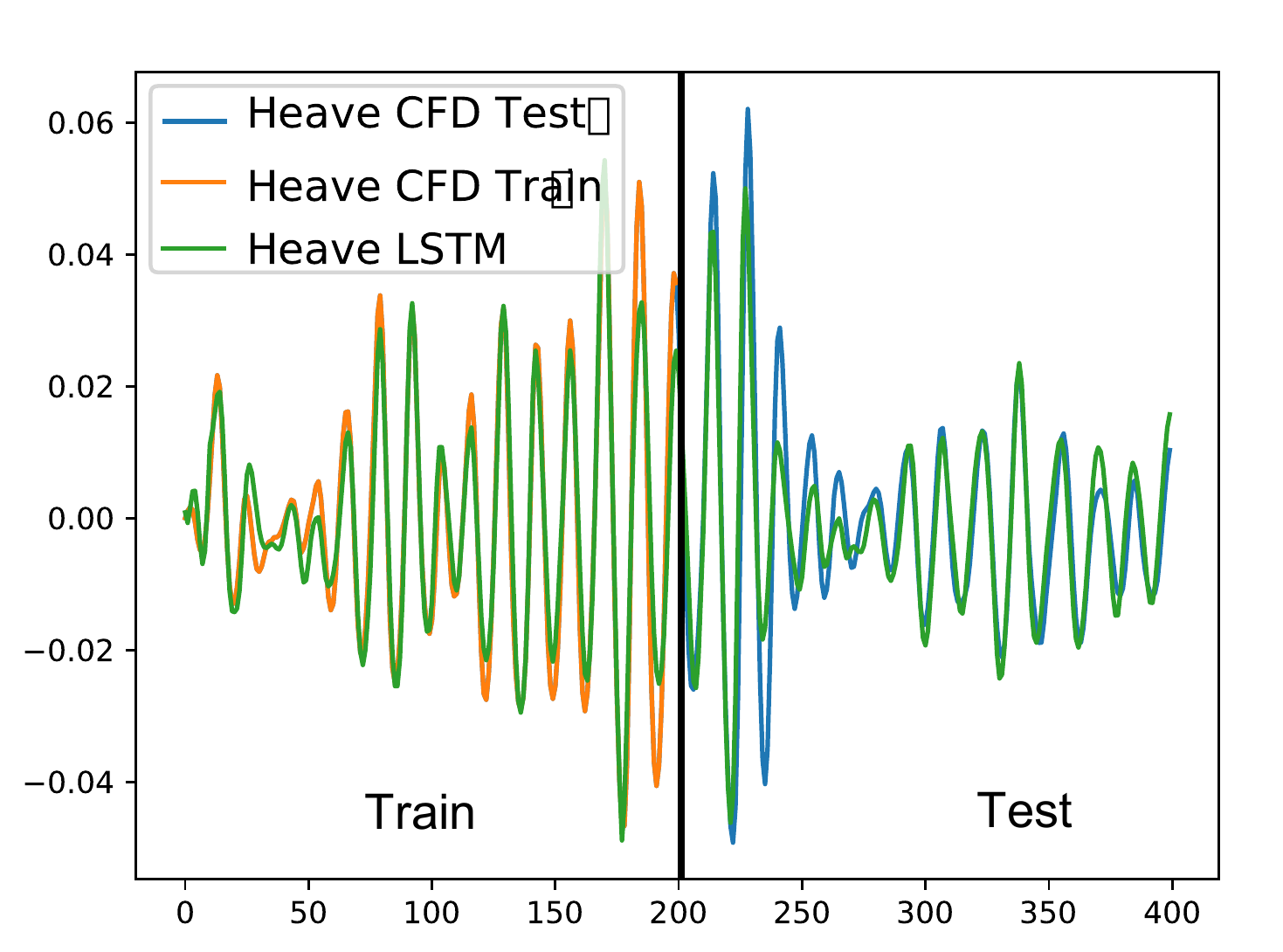}
        \caption{\scriptsize  RSE = 3.7029e-05.}
        \label{ConvIrrFig:b}
    \end{subfigure}%
     ~
    \begin{subfigure}[t]{0.32\textwidth}
        \centering
        \includegraphics[width=\columnwidth]{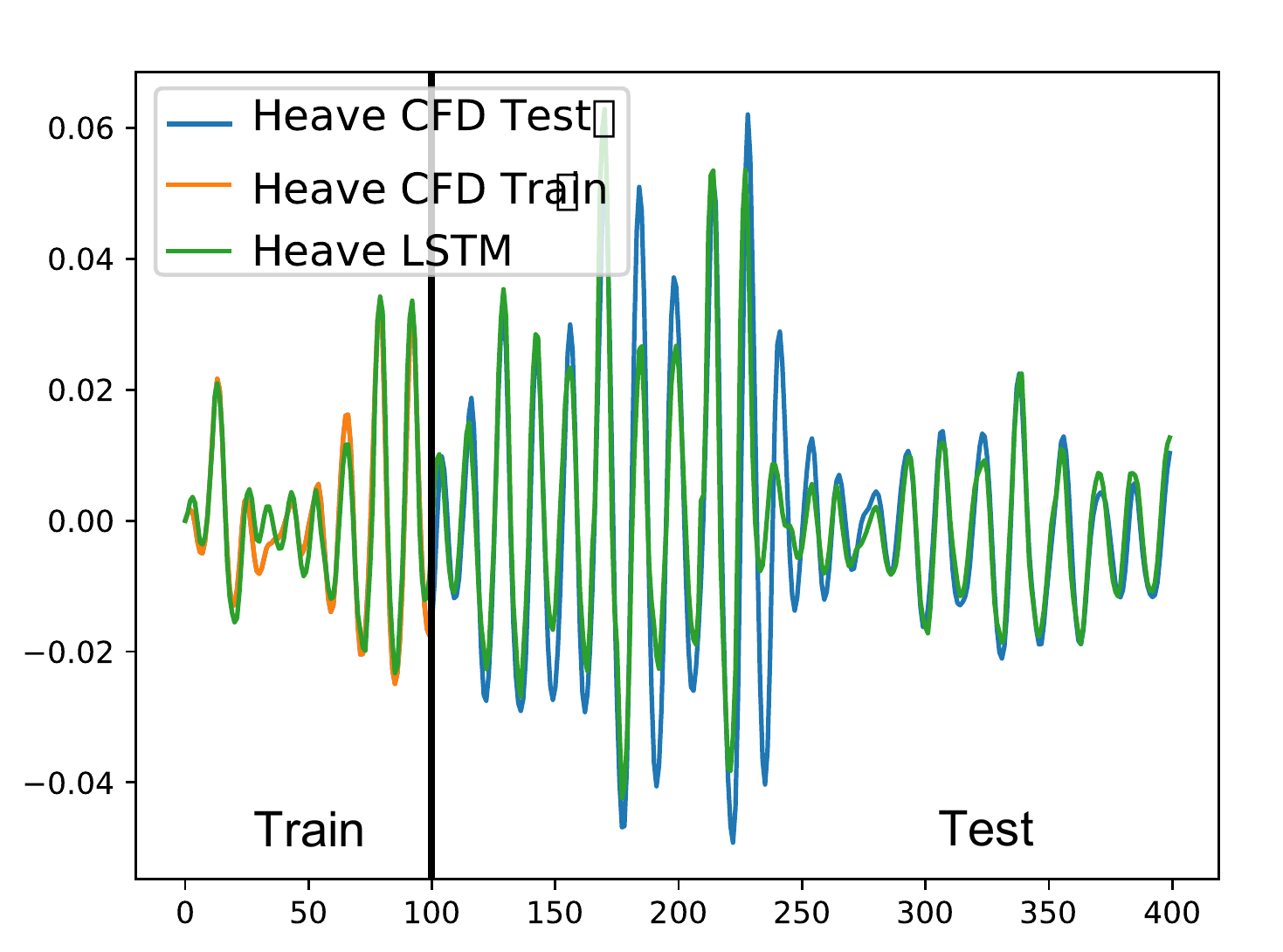}
        \caption{\scriptsize  RSE = 5.7327e-05.}
        \label{ConvIrrFig:c}
    \end{subfigure}%
    
 %%%%%%%%%%%%%%%%%%%%%%%%%%%%%%%%%%%%%%%%%%%%%%%%%%%%%%%%%%%%%%%%%%%%%%%%%%%%%%%%%%%%%%%%%%%%%%%%%
    \begin{subfigure}[t]{0.32\textwidth}
        \centering
        \includegraphics[width=\columnwidth]{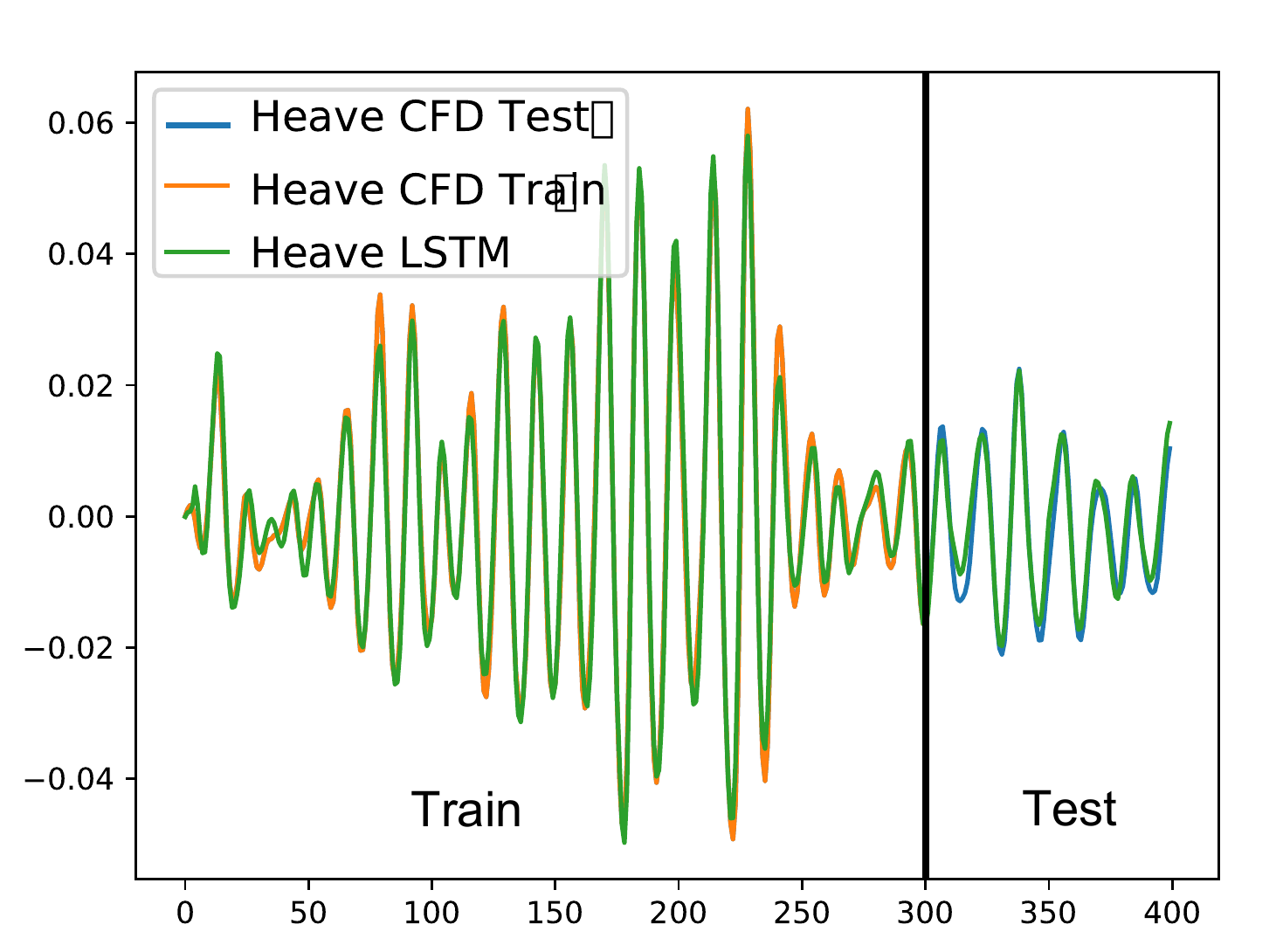}
        \caption{\scriptsize RSE = 8.4781e-06.} 
		\label{ConvIrrFig:d}    
    \end{subfigure}
    ~
    \begin{subfigure}[t]{0.32\textwidth}
        \centering
        \includegraphics[width=\columnwidth]{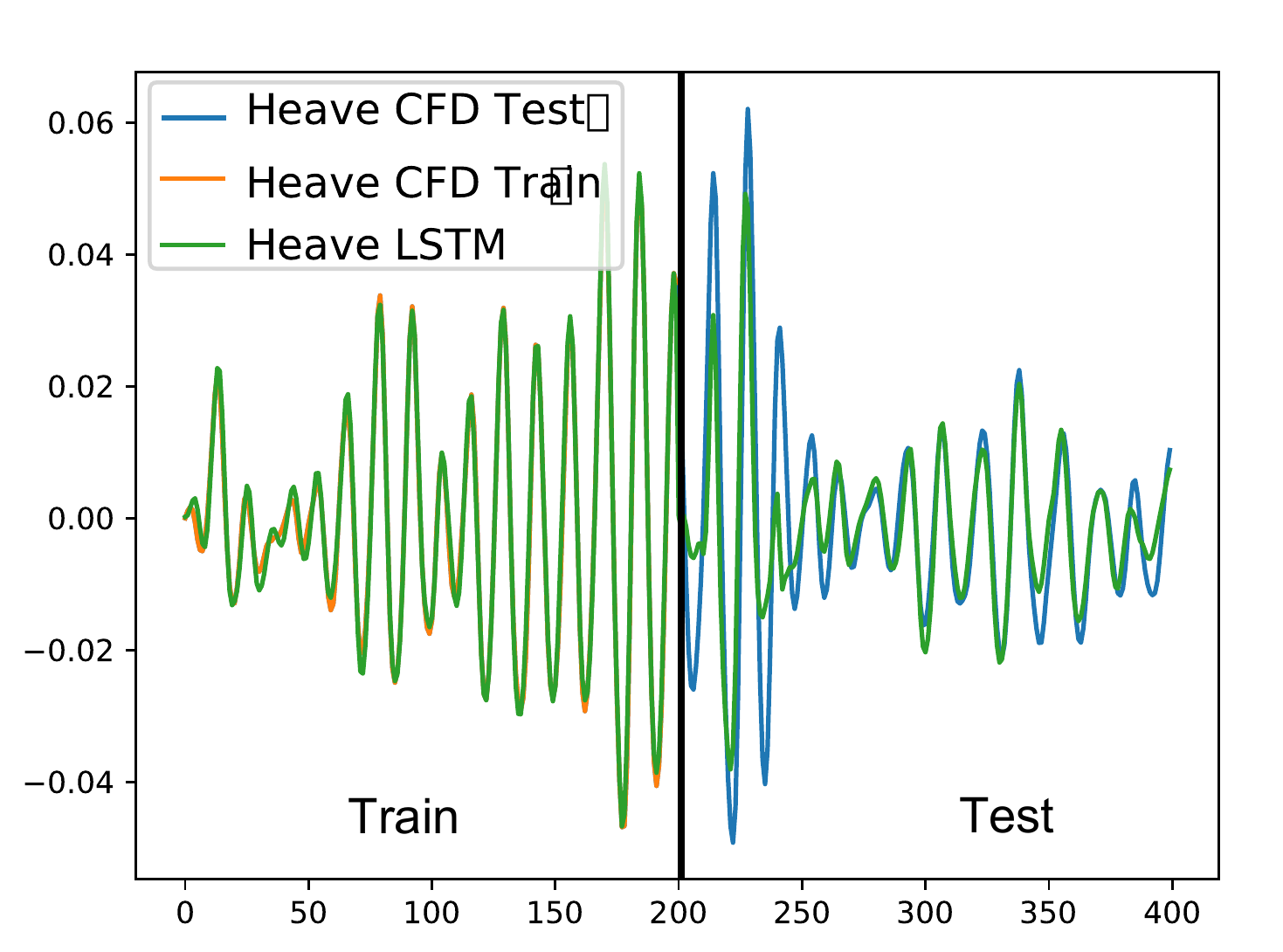}
        \caption{\scriptsize RSE = 4.5099e-05.}
        \label{ConvIrrFig:e}
    \end{subfigure}%
     ~
    \begin{subfigure}[t]{0.32\textwidth}
        \centering
        \includegraphics[width=\columnwidth]{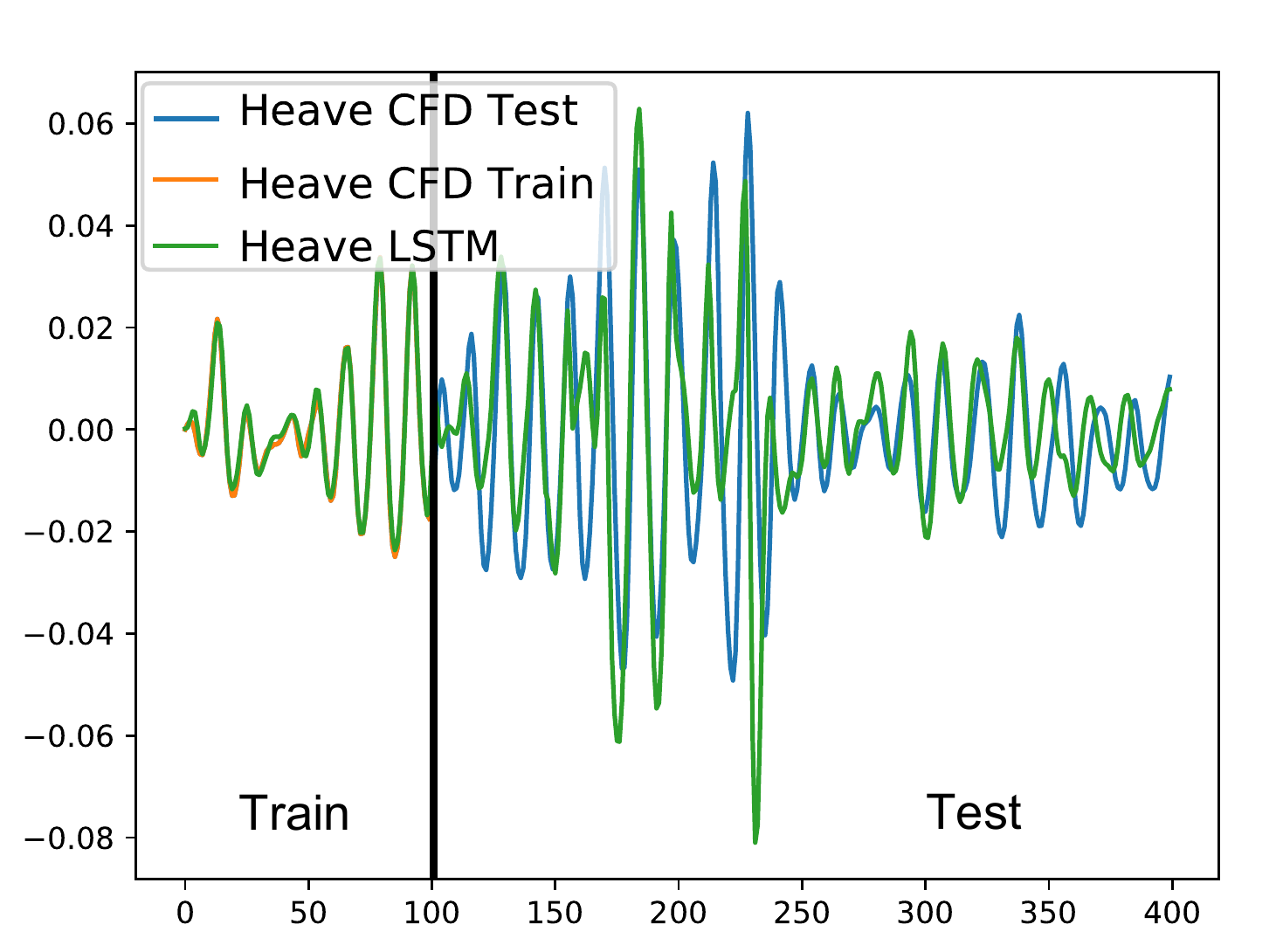}
        \caption{\scriptsize RSE = 2.5656e-04.}
        \label{ConvIrrFig:f}
    \end{subfigure}%
%%%%%%%%%%%%%%%%%%%%%%%%%%%%%%%%%%%%%%%%%%%%%%%%%%%%%%%%%%%%%%%%%%%%%%%%%%%%%%%%%%%%%%%%%%%%%%%%%%%

    \caption{\textit{Vertical motion of the catamaran vessel using two LSTM networks with 1 layer and 15 neurons (upper row) and 3 layers and 15 neurons (lower row).} The left column corresponds to $2/3$ training data from \cref{fig:Inputc}, the middle column corresponds to $1/2$ of the training data, and the right column corresponds to $1/4$ of the training data. Each time step corresponds to $\Delta t=0.0625s$.}
\label{fig:ConvIrreg}
\end{figure*}

\begin{figure*}[] 
 \centering
 %%%%%%%%%%%%%%%%%%%%%%%%%%%%%%%%%%%%%%%%%%%%%%%%%%%%%%%%%%%%%%%%%%%%%%%%%%%%%%%%%%%%%%%%%%%%%%%%%
   
    \begin{subfigure}[t]{0.32\textwidth}
        \centering
        \includegraphics[width=\columnwidth]{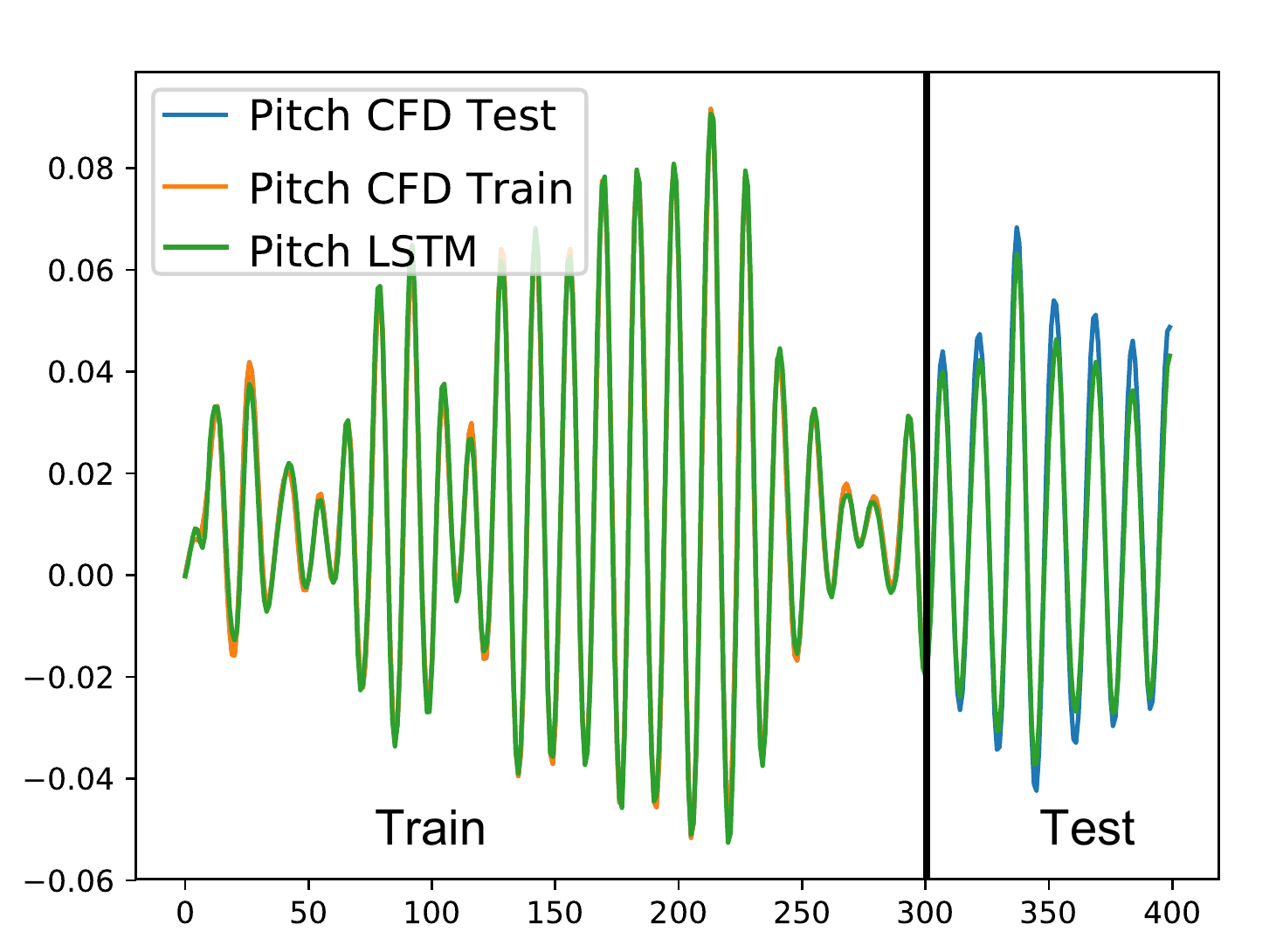}
        \caption{\scriptsize RSE =1.0249e-05.} 
		\label{ConvIrrFig:g}    
    \end{subfigure}
    ~
    \begin{subfigure}[t]{0.32\textwidth}
        \centering
        \includegraphics[width=\columnwidth]{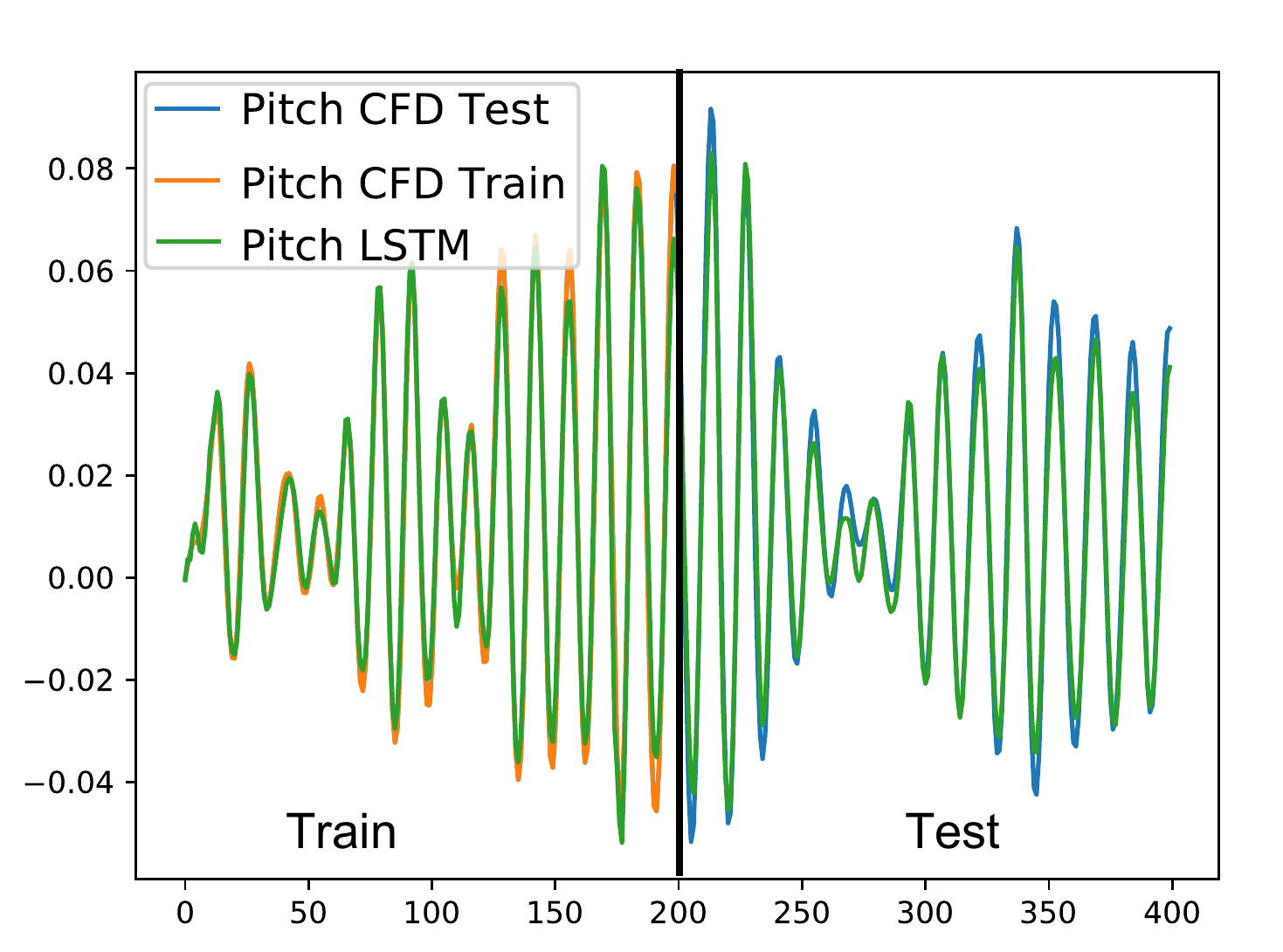}
        \caption{\scriptsize RSE =2.3605e-05.}
        \label{ConvIrrFig:h}
    \end{subfigure}%
     ~
    \begin{subfigure}[t]{0.32\textwidth}
        \centering
        \includegraphics[width=\columnwidth]{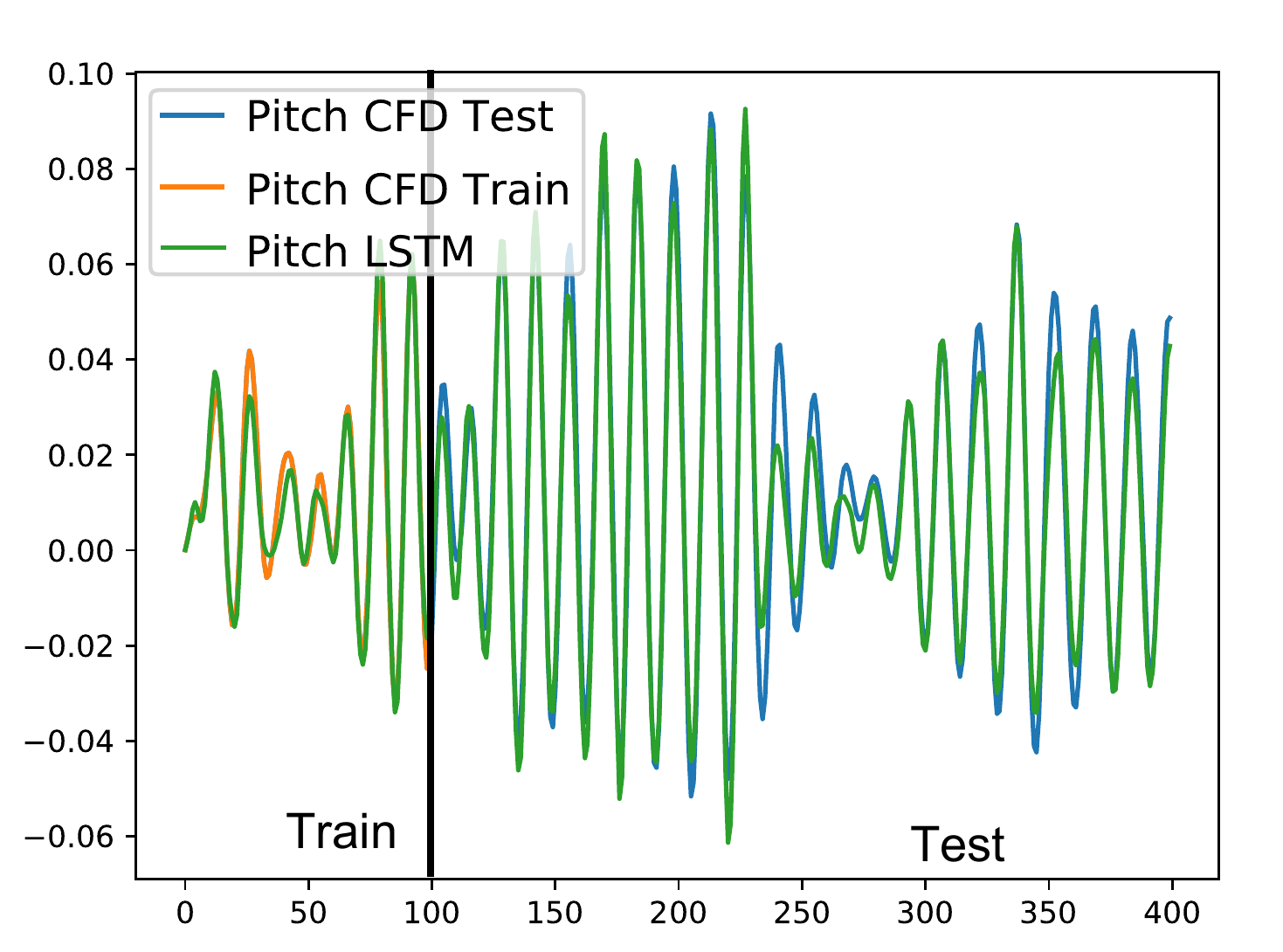}
        \caption{\scriptsize RSE =4.0008e-05.}
        \label{ConvIrrFig:i}
    \end{subfigure}%
 %%%%%%%%%%%%%%%%%%%%%%%%%%%%%%%%%%%%%%%%%%%%%%%%%%%%%%%%%%%%%%%%%%%%%%%%%%%%%%%%%%%%%%%%%%%%%%%%%
 
    \begin{subfigure}[t]{0.32\textwidth}
        \centering
        \includegraphics[width=\columnwidth]{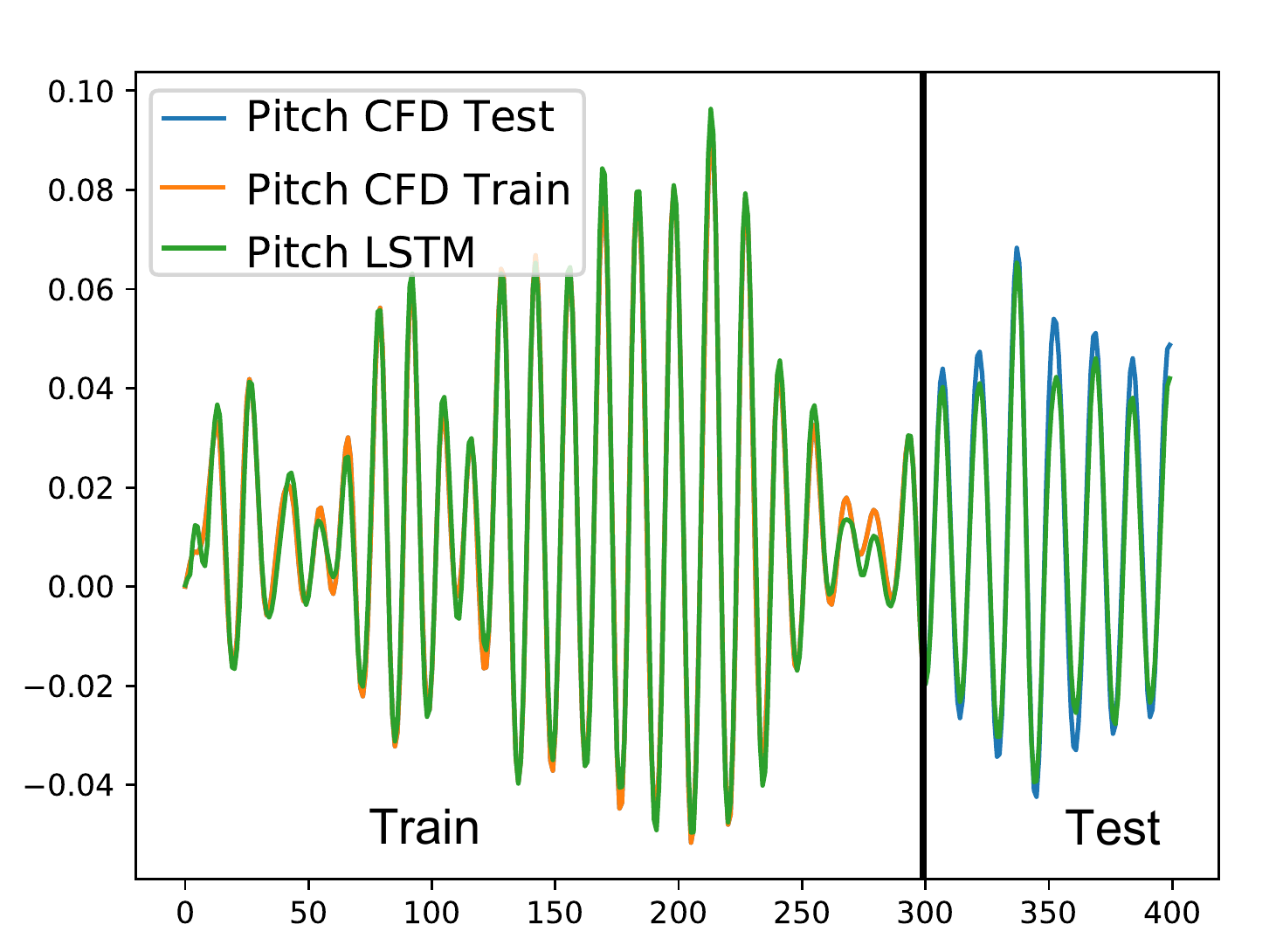}
        \caption{\scriptsize RSE = 1.2446e-05.} 
		\label{ConvIrrFig:j}    
    \end{subfigure}
    ~
    \begin{subfigure}[t]{0.32\textwidth}
        \centering
        \includegraphics[width=\columnwidth]{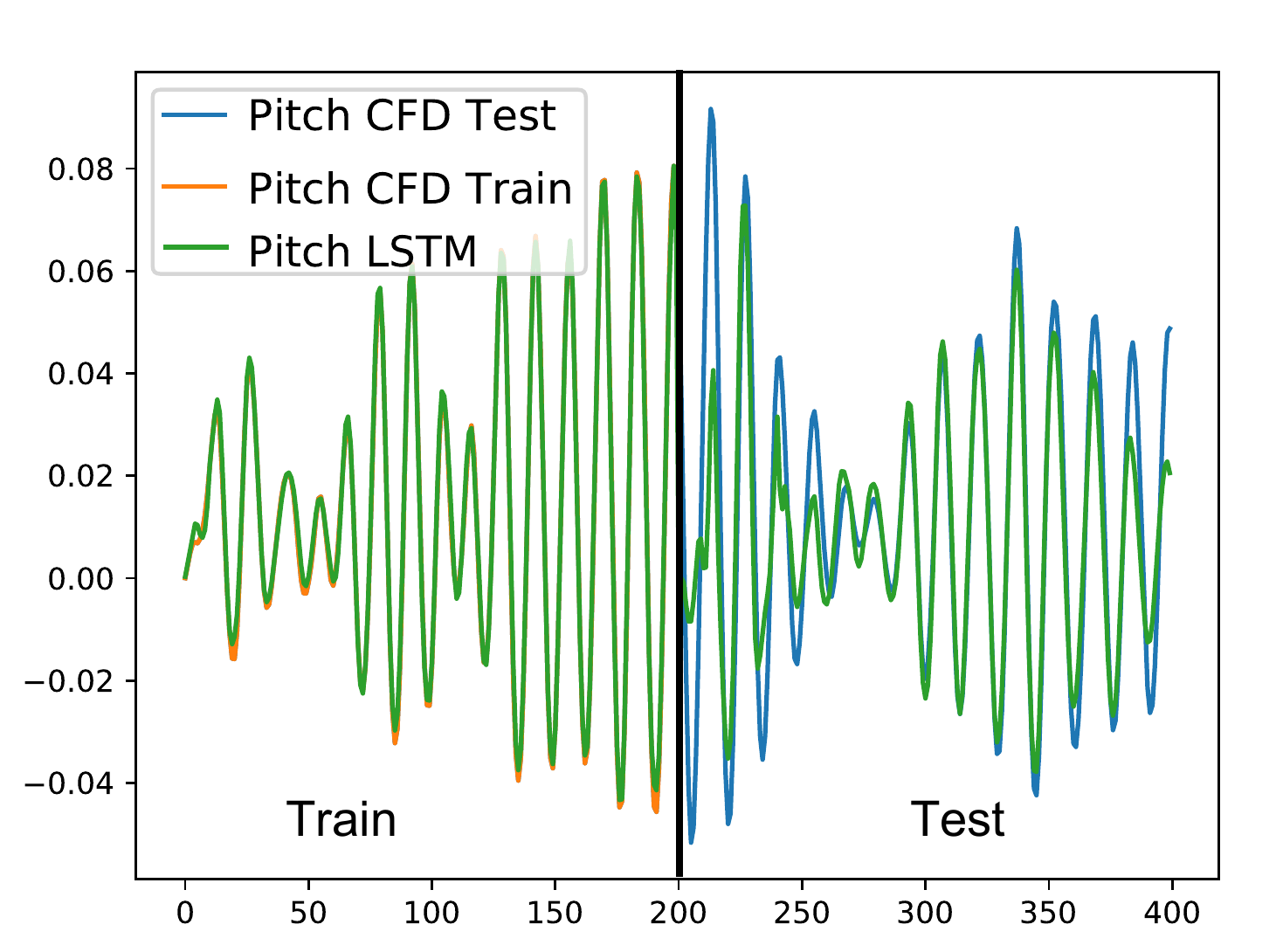}
        \caption{\scriptsize RSE = 1.1840e-04.}
        \label{ConvIrrFig:k}
    \end{subfigure}%
     ~
    \begin{subfigure}[t]{0.32\textwidth}
        \centering
        \includegraphics[width=\columnwidth]{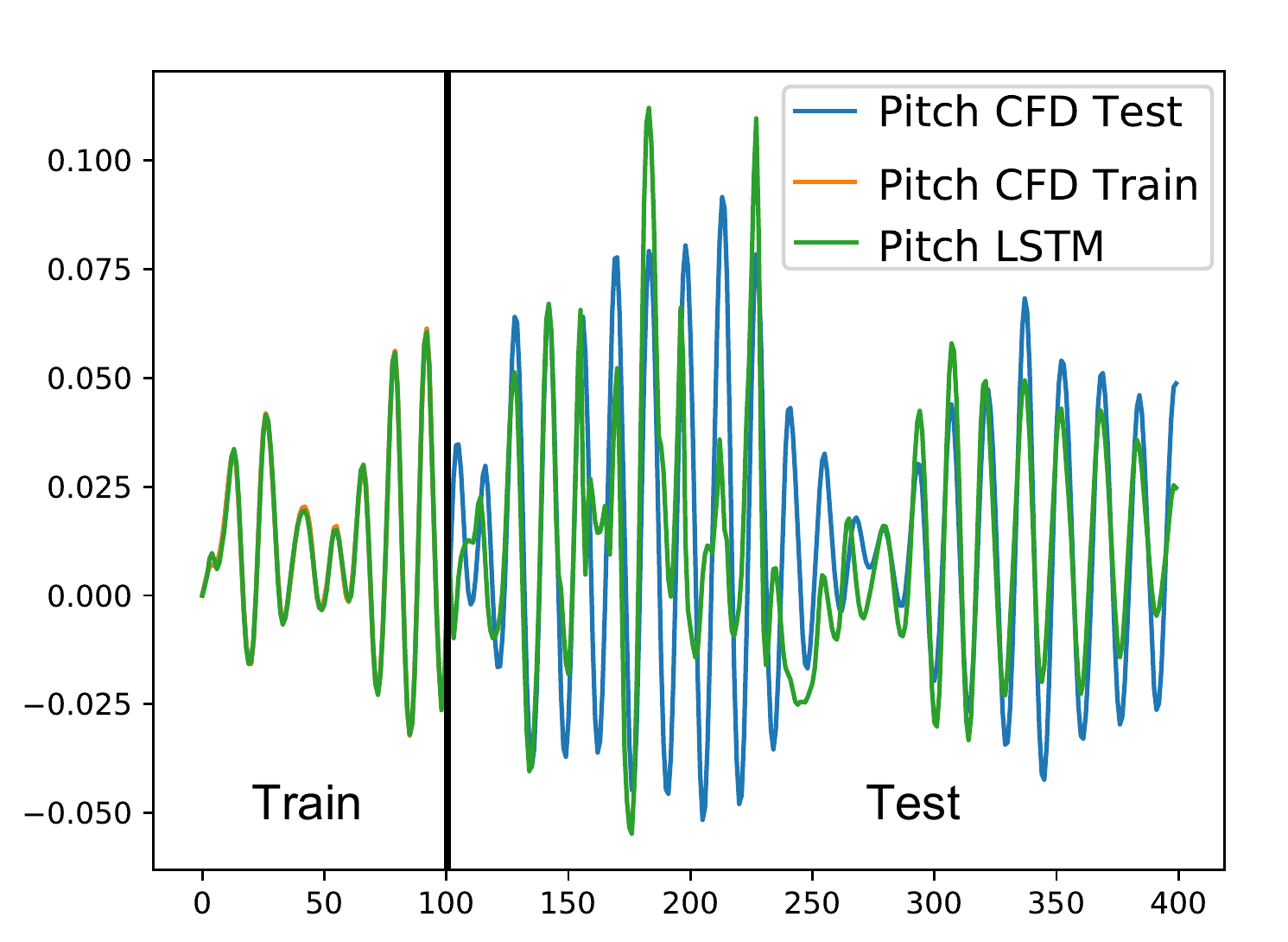}
        \caption{\scriptsize RSE = 4.3936e-04.}
        \label{ConvIrrFig:l}
    \end{subfigure}%
    \caption{\textit{Angular motion of the catamaran vessel using two LSTM networks with 1 layer and 15 neurons (upper row) and 3 layers and 15 neurons (lower row).} The left column corresponds to $2/3$ training data from \cref{fig:Inputc}, the middle column corresponds to $1/2$ of the training data, and the right column corresponds to $1/4$ of the training data. Each time step corresponds to $\Delta t=0.0625s$.}
\label{fig:ConvIrreg2}
\end{figure*}

\begin{figure*}[] 
 \centering
 %%%%%%%%%%%%%%%%%%%%%%%%%%%%%%%%%%%%%%%%%%%%%%%%%%%%%%%%%%%%%%%%%%%%%%%%%%%%%%%%%%%%%%%%%%%%%%%%%
    \begin{subfigure}[t]{0.95\textwidth}
        \centering
        \includegraphics[width=\columnwidth]{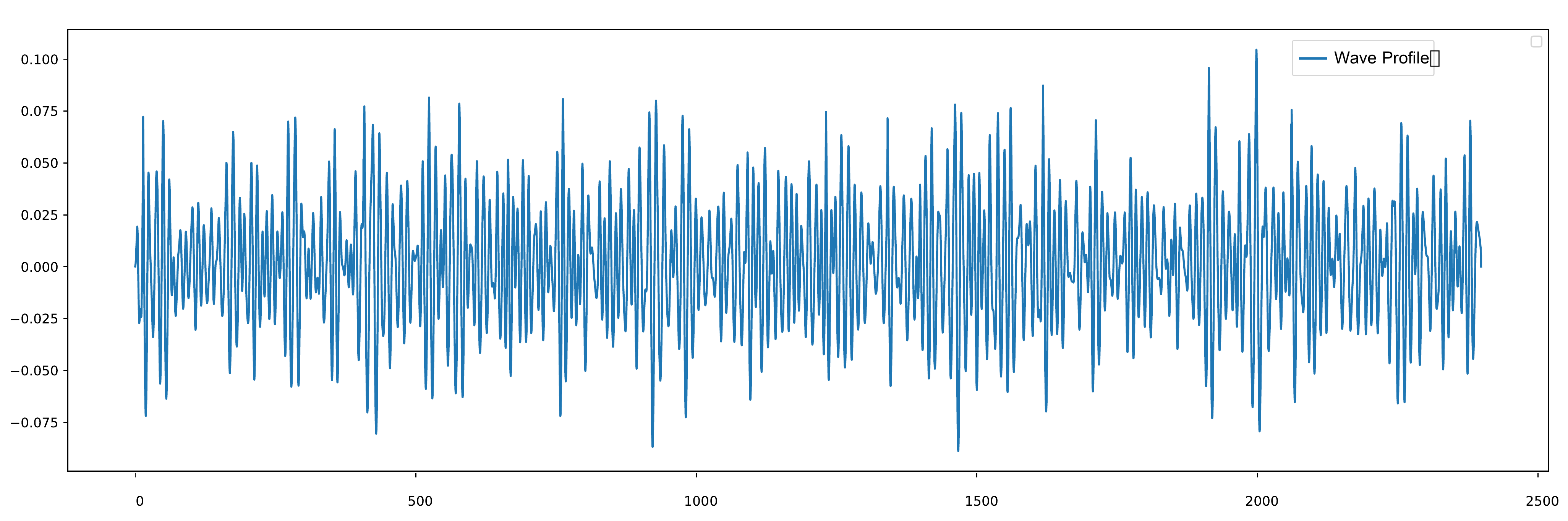}
        \caption{}
        \label{LongConvIrrFig:a}
    \end{subfigure}% 
    
    \begin{subfigure}[t]{0.95\textwidth}
        \centering
        \includegraphics[width=\columnwidth]{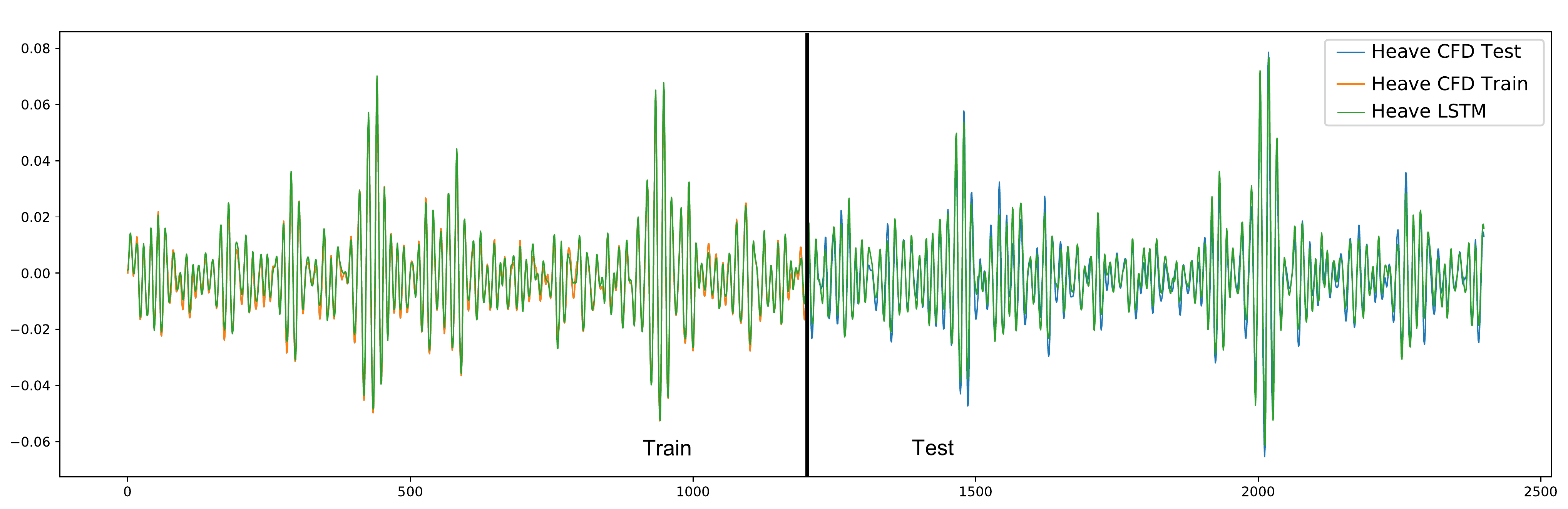}
        \caption{}
        \label{LongConvIrrFig:b}
    \end{subfigure}%
     
    \begin{subfigure}[t]{0.95\textwidth}
        \centering
        \includegraphics[width=\columnwidth]{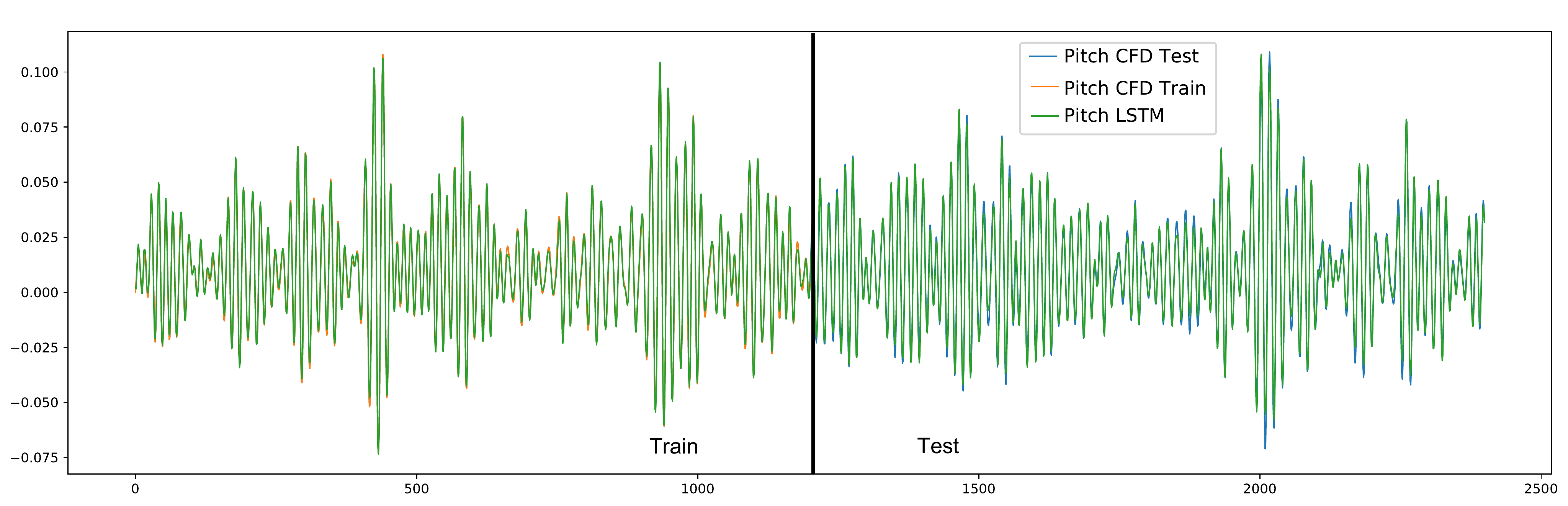}
        \caption{}
        \label{LongConvIrrFig:b}
    \end{subfigure}%

 %%%%%%%%%%%%%%%%%%%%%%%%%%%%%%%%%%%%%%%%%%%%%%%%%%%%%%%%%%%%%%%%%%%%%%%%%%%%%%%%%%%%%%%%%%%%%%%%%

    \caption{\textit{Long-time predictions of vertical (b) and angular (c) motions of the catamaran vessel using a LSTM network with 2 layers and 10 neurons.} Figure (a) shows the surface elevation input to the network. The vertical line in (b),(c) denotes the beginning of testing. Each time step corresponds to $\Delta t=0.0625s$.}
\label{fig:ConvIrregLong}
\end{figure*}

The inputs used to train and test the networks are shown in \cref{fig:InputIrreg}. Results of this study are presented in \cref{fig:Irreg,fig:ConvIrreg,fig:ConvIrreg2,fig:ConvIrregLong}. In particular, in \cref{fig:Irreg} we plot the heave and pitch motions predicted by a LSTM network with 20 neurons and 1 layer. On the left column we present predictions based on training data and on the right column we present predictions based on unseen data. The training data set includes 3 different sequences for the surface elevation, similar to the one shown in \cref{fig:Inputa}. The testing predictions (right column of figure 6) are based on surface elevation inputs shown in \cref{fig:Inputb}. We observe that this shallow network is adequate in approximating the two DOF catamaran vessel subject to irregular waves. 

A more systematic investigation of the network architectures and the amount of training is presented in figures 7 and 8. These figures represent results from a  set of 64 parametric variations of five network architecture parameters. These parameters include neurons, hidden layers, training steps, number of sequences in the training process, and the fraction of anyone sequence used in the training process. In the first 6 plots, sub-figures \ref{ConvIrrFig:a} - \ref{ConvIrrFig:f}, we analyze the evolution of motion predictions when we vary the fraction of  sequence used in the training process (\ref{ConvIrrFig:a} $\rightarrow$ \ref{ConvIrrFig:c} or \ref{ConvIrrFig:d} $\rightarrow$ \ref{ConvIrrFig:f} ) or when we vary the number of hidden layers (\ref{ConvIrrFig:a} $\rightarrow$ \ref{ConvIrrFig:d} or \ref{ConvIrrFig:b} $\rightarrow$ \ref{ConvIrrFig:e} or \ref{ConvIrrFig:c} $\rightarrow$ \ref{ConvIrrFig:f}). The second set in \cref{fig:ConvIrreg2} of subplots (\ref{ConvIrrFig:g} - \ref{ConvIrrFig:l}) is similar to the first set, but the only change is the variable in the plots, i.e.,  instead of the vertical motion (heave) we plot the angular motion (pitch). It is remarkable that even with a small training data set accurate predictions of both heave and pitch are obtained as long
as we adjust the LSTM architecture to avoid overfitting.
The accuracy of long-term LSTM predictions is studied in \cref{fig:ConvIrregLong}b-c. The LSTM network inputs are presented in \cref{LongConvIrrFig:a}. The network (with 2 layers and 10 neurons) is trained on the first half of the sea surface elevation time series and forecasts the vertical and angular motions for the second half. We can see a very accurate and stable prediction of the motion dynamics even for long-term, which is in agreement with unused data from the CFD solver. 
%One reasonable explanation is the short memory of the system and the fact that, given that the vessel is still floating it must remain in the proximity of the free surface.

In summary, we have observed the following main trends:

\begin{enumerate}
\item
\textit As we decrease the amount of information given to the network (compare column \ref{ConvIrrFig:a} $\downarrow$ \ref{ConvIrrFig:j} to \ref{ConvIrrFig:c} $\downarrow$ \ref{ConvIrrFig:l}) we can clearly see that the network is less accurate and is more prone to over-fitting. 
\item
\textit Over-fitting is emphasized in subplots \ref{ConvIrrFig:f} and \ref{ConvIrrFig:l}, which is natural since they are the cases with the greatest number of parameters, because they have 3 hidden layers. So as we increase the expressive power of the network we must also increase the amount of information given for training.
\item
\textit LSTM networks provide a stable long term predictor in terms of amplitude, frequency and phase of the vessel motions.
\end{enumerate}

Our conclusions regarding our studies with respect to the network architecture convergence are as follows:

\begin{enumerate}
\item
Given the amount of data available, 1 hidden layer with 15 neurons approximates accurately the functional of the catamaran motions for a mild sea state.
\item
Given enough data these predictions can be improved with more expressive networks, however, these networks are also quite prone to overfitting. This can be concluded since accuracy in \cref{ConvIrrFig:d,ConvIrrFig:j} is better or equal than in \cref{ConvIrrFig:a,ConvIrrFig:g}; however, it is clearly worse when we reduce the amount of information in the training cases. To appreciate this, one can see that the accuracy in \cref{ConvIrrFig:f,ConvIrrFig:l} is worse than in \cref{ConvIrrFig:c,ConvIrrFig:i}.    
\item
The data modeling framework is both feasible and practical. It is also very fast. Specifically, it took about 120h (on a 20 processor computer) to run the CFD solver to obtain the training labeled data. To train the LSTM requires approximately half an hour on a GPU, while to obtain the predictions in our testing experiments requires only a fraction of a second. For example in \cref{ConvIrrFig:c,ConvIrrFig:i} we can train the network on the first $1/4$ of the sequence and reproduce the remaining $3/4$, at a negligible cost compared to using CFD for the entire sequence.
\end{enumerate}

\subsection{DTBM Vessel}

To further test the capabilities of the LSTM network we introduce a third DOF and we aim to model the motions of a second vessel in a severe sea state. In addition to heave and pitch, the vessel is now allowed to rotate around its longitudinal axis (rolling). This type of motion is affected strongly by the viscous effects, in contrast to heave and pitch motions that can be well approximated given the stiffness and mass matrix of the vessel. Nevertheless, it is also true that in the extreme sea state we consider in this case (sea state 8) many of the assumptions of the linear theory that is conventionally used do not hold. The nonlinearities in heave and pitch can be successfully modeled with nonlinear Boundary Element Methods (BEMs) that track the vessels' wet surface \cite{Kring2004}. However, BEM does not have the fidelity to model accurately roll motions, since the solver needs to resolve the  viscous terms of the Navier-Stokes equations to have the necessary information to accurately compute the roll damping of the motions. A validated approximation is given by the unsteady Reynolds average Navier-Stokes equations (URANS) methodology \cite{itemreference3}. However, given the complexity of this wave-structure interaction problem this requires a prohibitive expensive series of simulations from which we obtain data that is difficult to generalize, thus it is better to develop a surrogate model in the form of learning a functional to approximate the 3DOF for the DTBM vessel.

\begin{figure*}[] 
 \centering

 %%%%%%%%%%%%%%%%%%%%%%%%%%%%%%%%%%%%%%%%%%%%%%%%%%%%%%%%%%%%%%%%%%%%%%%%%%%%%%%%%%%%%%%%%%%%%%%%%
  
    \begin{subfigure}[t]{0.65\textwidth}
        \centering
        \includegraphics[width=\columnwidth]{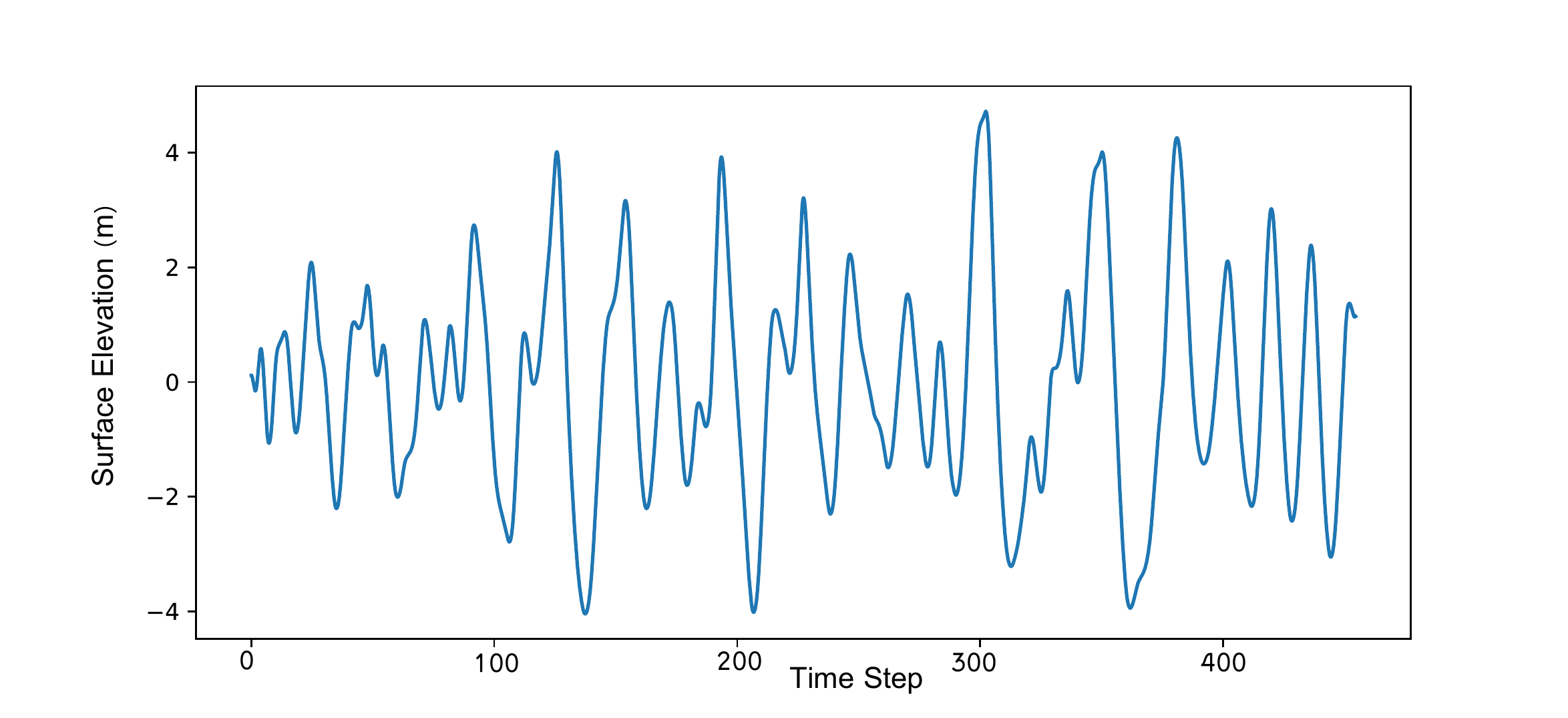}
        \caption{\scriptsize Profile of the wave in the longitudinal direction corresponding to {test} case one.} 
		\label{fig:j}    
    \end{subfigure}

    \begin{subfigure}[t]{0.65\textwidth}
        \centering
        \includegraphics[width=\columnwidth]{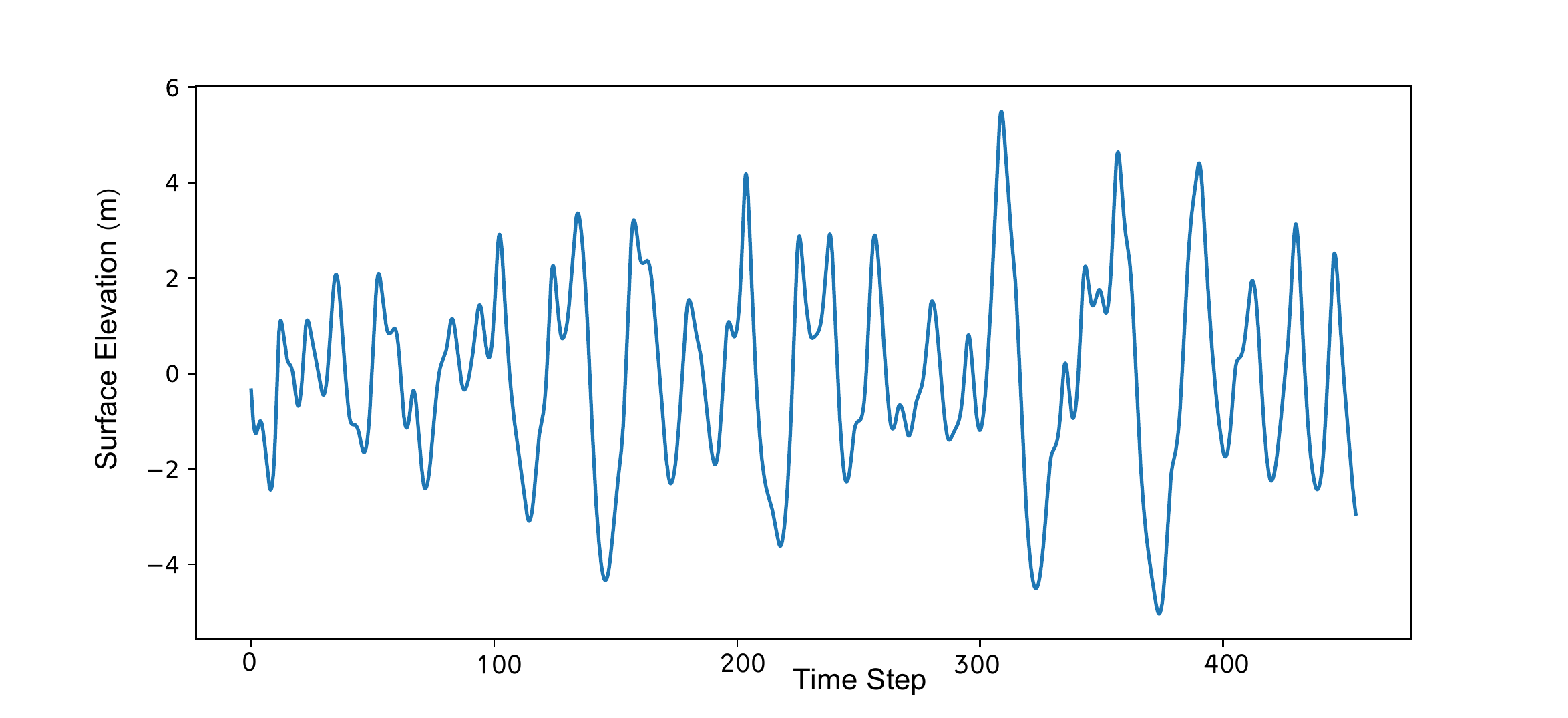}
        \caption{\scriptsize Profile of the wave in the transversal direction corresponding to {test} case one.} 
		\label{fig:k}    
    \end{subfigure}
       
 %%%%%%%%%%%%%%%%%%%%%%%%%%%%%%%%%%%%%%%%%%%%%%%%%%%%%%%%%%%%%%%%%%%%%%%%%%%%%%%%%%%%%%%%%%%%%%%%%
     \begin{subfigure}[t]{0.65\textwidth}
        \centering
        \includegraphics[width=\columnwidth]{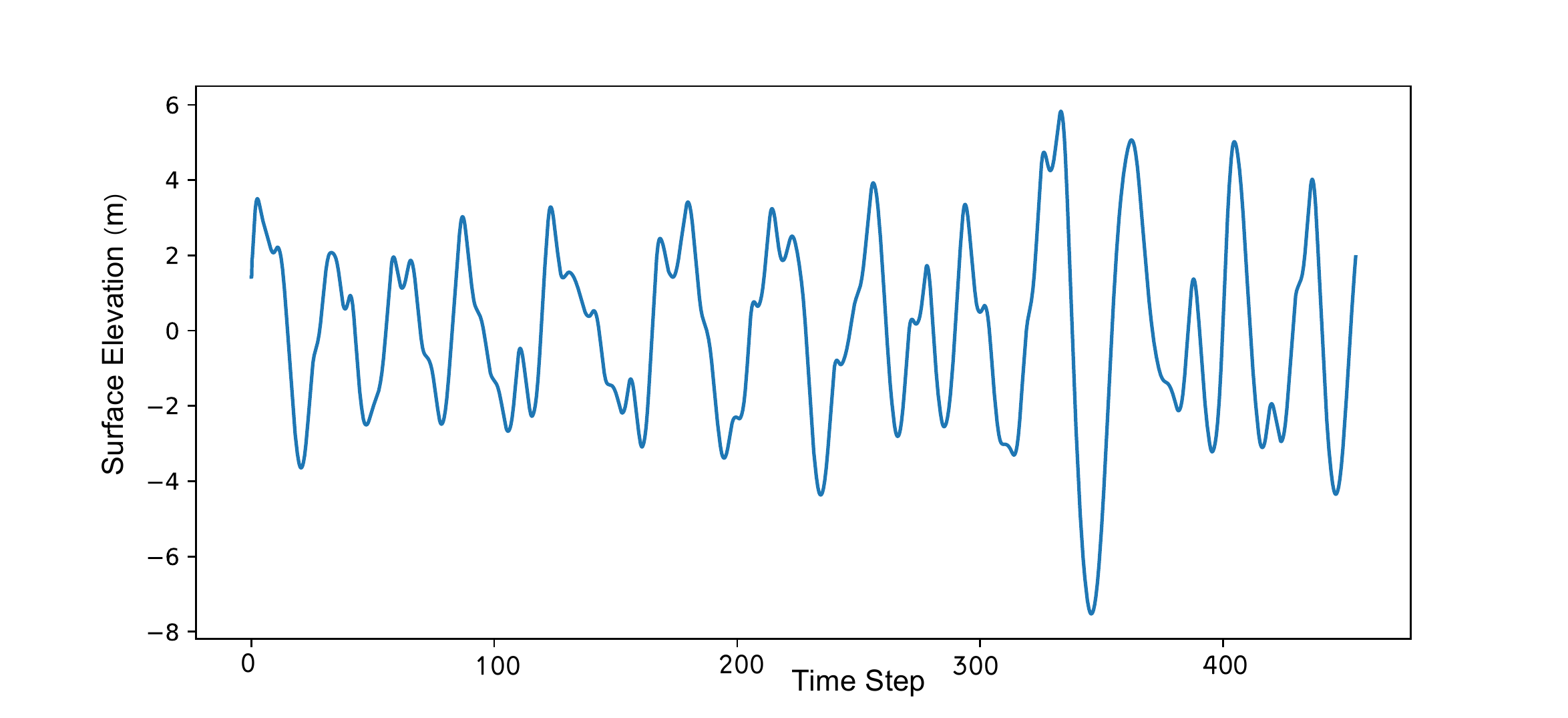}
        
        \caption{\scriptsize Profile of the wave in the longitudinal direction corresponding to {test} case two.} 
		\label{fig:j}    
    \end{subfigure}

    \begin{subfigure}[t]{0.65\textwidth}
        \centering
        \includegraphics[width=\columnwidth]{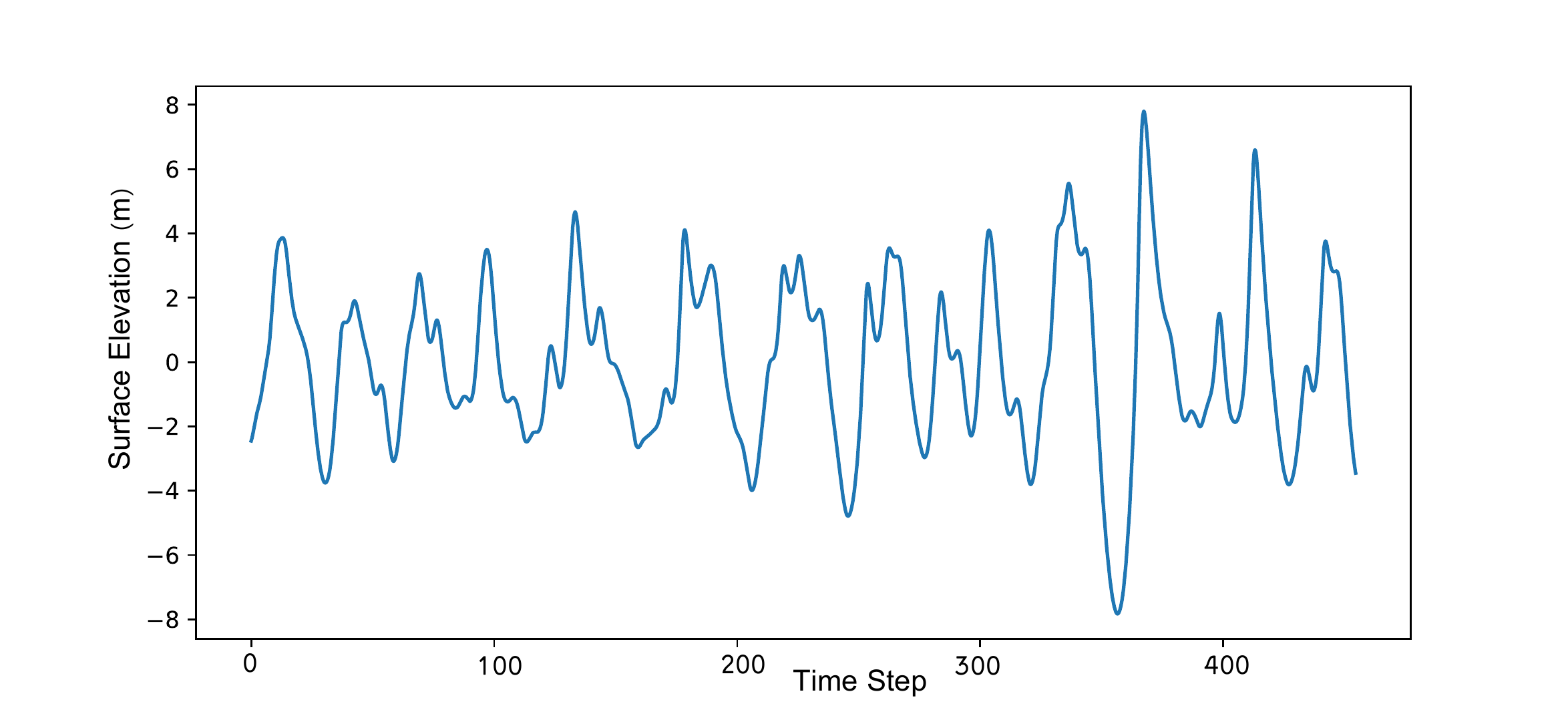}
        \caption{\scriptsize Profile of the wave in the transversal direction corresponding to {test} case two.} 
		\label{fig:k}    
    \end{subfigure}
 %%%%%%%%%%%%%%%%%%%%%%%%%%%%%%%%%%%%%%%%%%%%%%%%%%%%%%%%%%%%%%%%%%%%%%%%%%%%%%%%%%%%%%%%%%%%%%%%%
    
    \medskip
    \caption{\textit{Network surface elevation inputs for the DTMB vessel}. They are two-dimensional and represent a long crested irregular oblique waves. Each time step is $\Delta t = 0.2s$.} 
\label{fig:WaveCuts}
\end{figure*}

\begin{figure*}[] 
 \centering

 %%%%%%%%%%%%%%%%%%%%%%%%%%%%%%%%%%%%%%%%%%%%%%%%%%%%%%%%%%%%%%%%%%%%%%%%%%%%%%%%%%%%%%%%%%%%%%%%%
 
    \begin{subfigure}[t]{0.4\textwidth}
        \centering
        \includegraphics[width=\columnwidth]{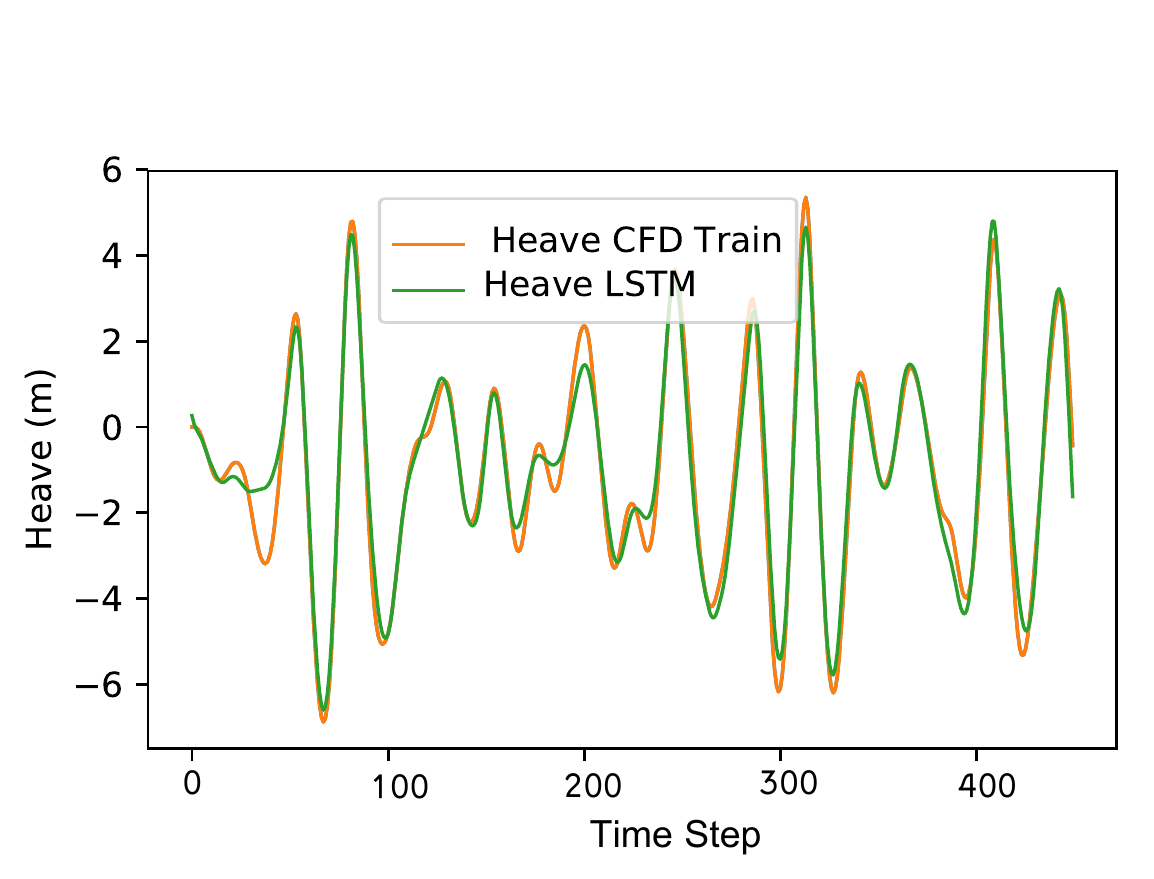}
        
        \caption{\scriptsize RSE = 0.0380} 
		\label{fig:j}    
    \end{subfigure}
~
    \begin{subfigure}[t]{0.4\textwidth}
        \centering
        \includegraphics[width=\columnwidth]{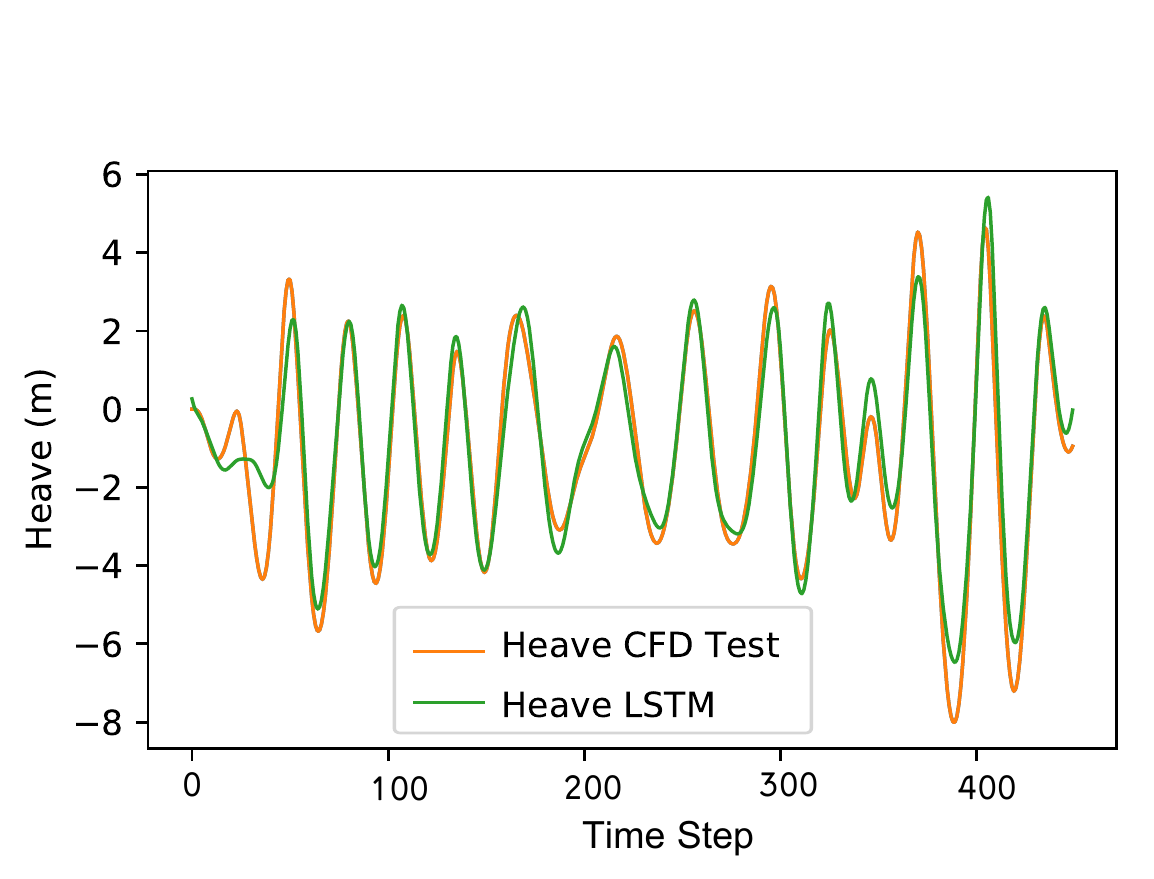}
        
        \caption{\scriptsize RSE = 0.0773} 
		\label{fig:k}    
    \end{subfigure}

%%%%%%%%%%%%%%%%%%%%%%%%%%%%%%%%%%%%%%%%%%%%%%%%%%%%%%%%%%%%%%%%%%%%%%%%%%%%%%%%%%%%%%%%%%%%%%%%%
 
     \begin{subfigure}[t]{0.4\textwidth}
        \centering
        \includegraphics[width=\columnwidth]{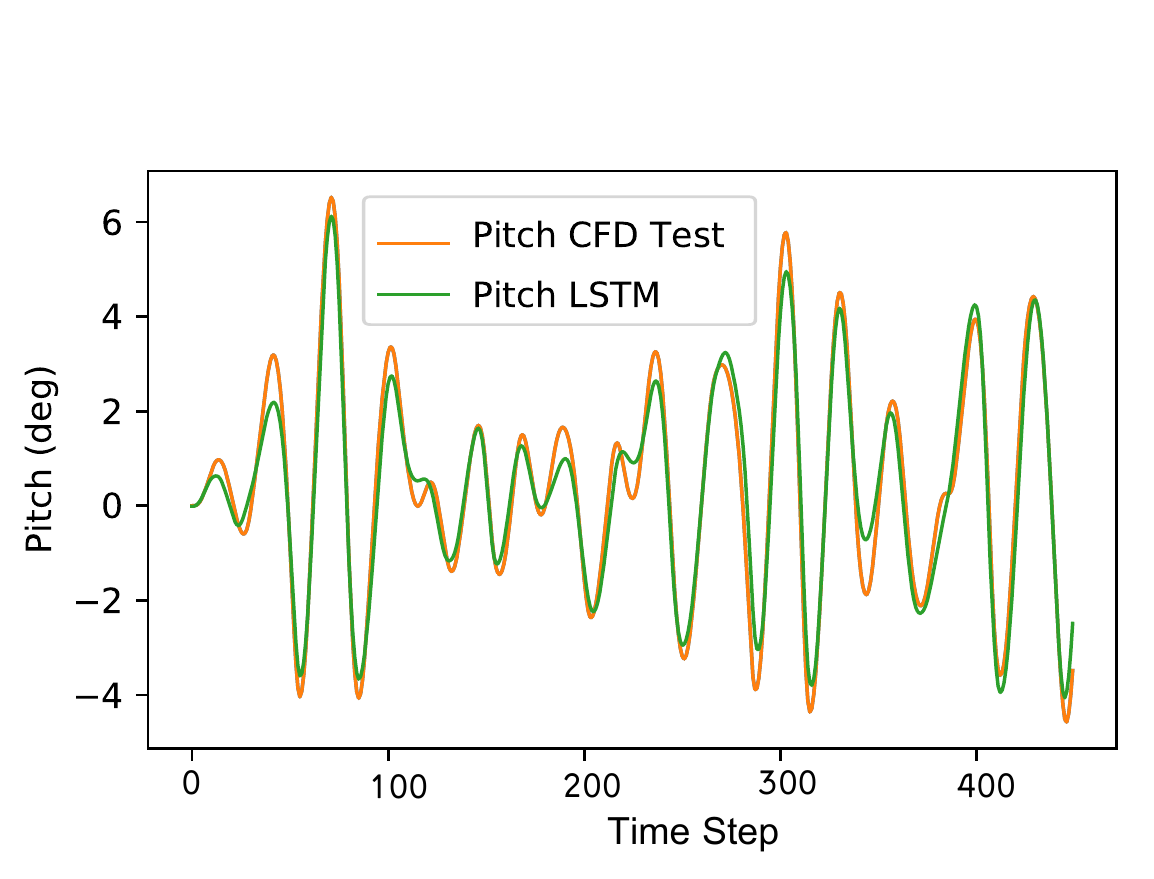}
        
        \caption{\scriptsize RSE = 0.0444} 
		\label{fig:j}    
    \end{subfigure}
~
    \begin{subfigure}[t]{0.4\textwidth}
        \centering
        \includegraphics[width=\columnwidth]{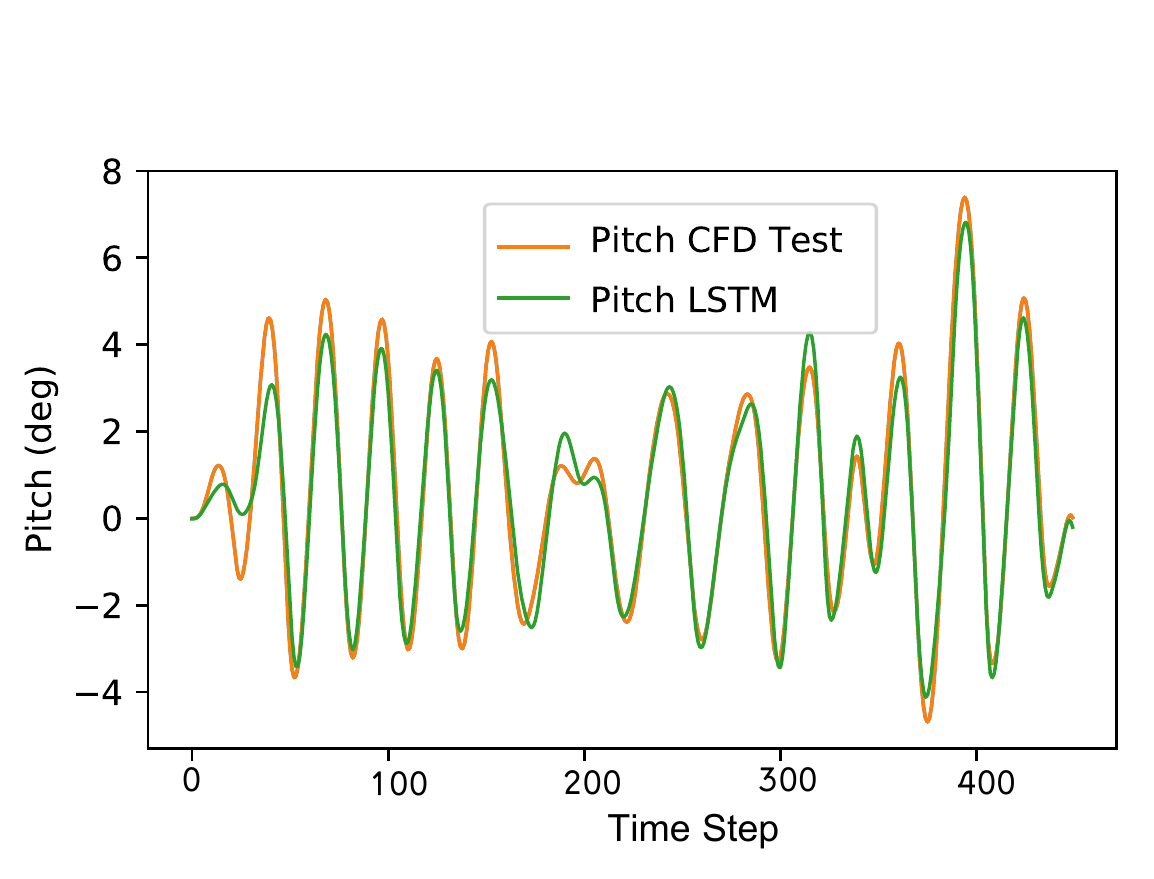}
        
        \caption{\scriptsize RSE = 0.0463} 
		\label{fig:k}    
    \end{subfigure}

%%%%%%%%%%%%%%%%%%%%%%%%%%%%%%%%%%%%%%%%%%%%%%%%%%%%%%%%%%%%%%%%%%%%%%%%%%%%%%%%%%%%%%%%%%%%%%%%%

    \begin{subfigure}[t]{0.4\textwidth}
        \centering
        \includegraphics[width=\columnwidth]{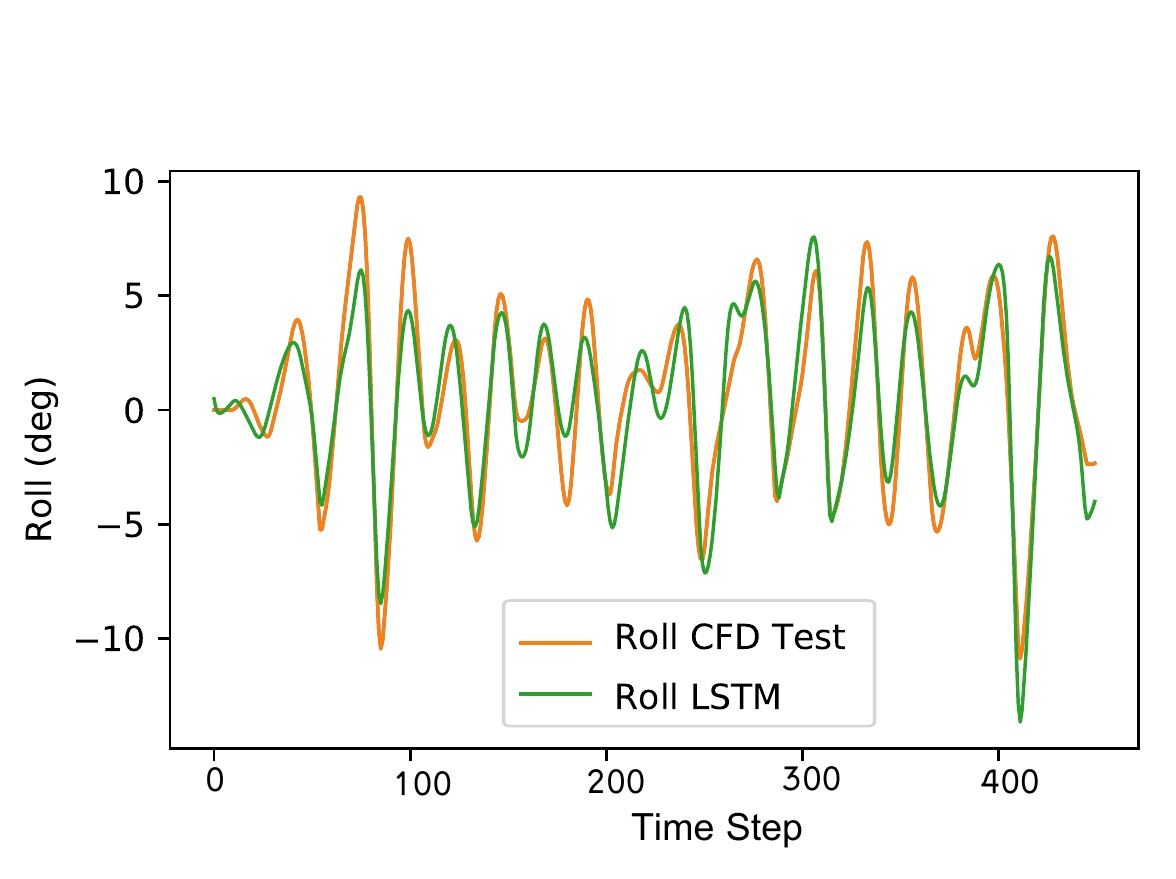}
        
        \caption{\scriptsize RSE = 0.1490} 
		\label{fig:j}    
    \end{subfigure}
~
    \begin{subfigure}[t]{0.4\textwidth}
        \centering
        \includegraphics[width=\columnwidth]{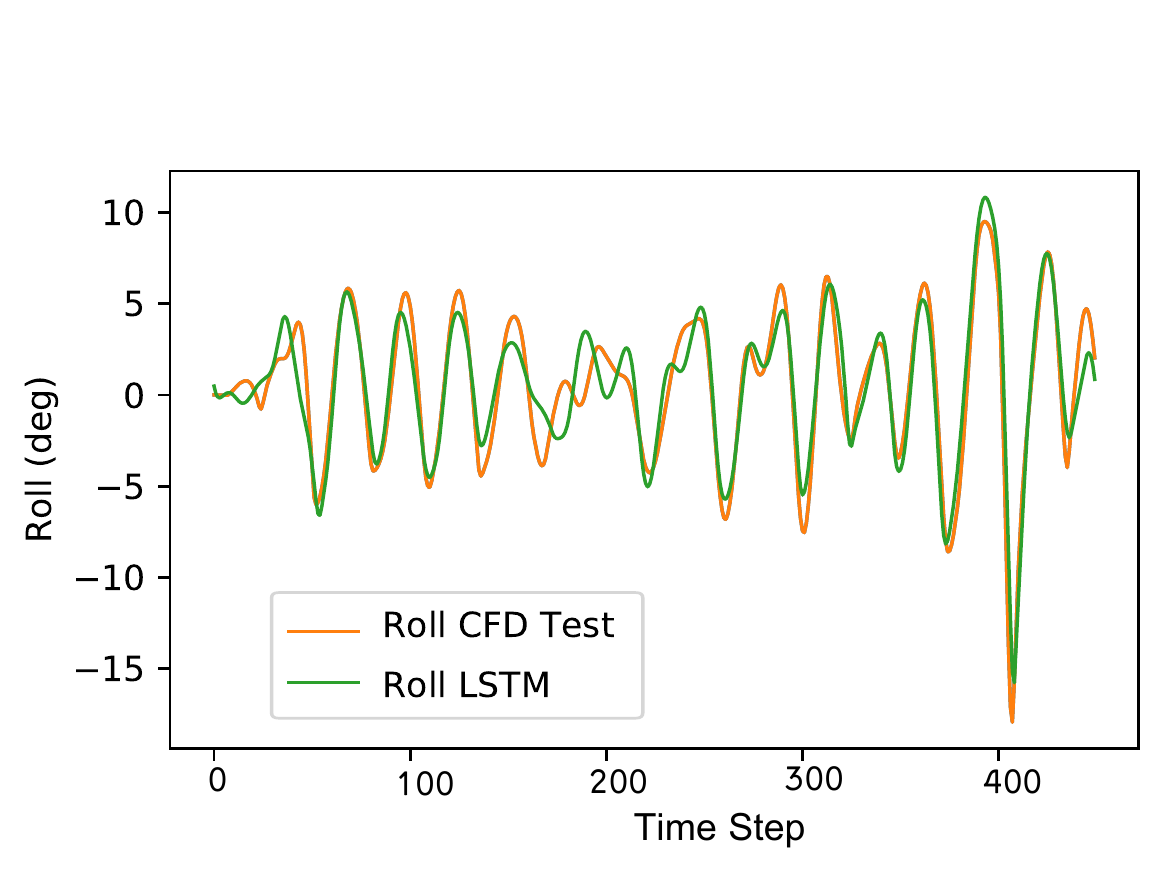}
        
        \caption{\scriptsize RSE = 0.1215} 
		\label{fig:k}    
    \end{subfigure}

    \begin{subfigure}[t]{\textwidth}
    \begin{minipage}{\textwidth}
    \centering
    \fontsize{7}{8}\selectfont
    \begin{tabular}{|c|c|c|c|c|c|c|c|c|}
    \hline
    \hline
    Network            & Case   & \begin{tabular}[c]{@{}c@{}}Heave \\ \\ RSE\end{tabular} & \begin{tabular}[c]{@{}c@{}}Pitch \\ \\ RSE\end{tabular} & \begin{tabular}[c]{@{}c@{}}Roll \\ \\ RSE\end{tabular} & \begin{tabular}[c]{@{}c@{}}Overall \\ \\ RSE\end{tabular} & Layers              & Neurons             & \begin{tabular}[c]{@{}c@{}}Train \\ \\ Steps\end{tabular} \\ \hline
    \multirow{2}{*}{1} & Test 1 & 0.0770                                                      & 0.0734                                                      & 0.1595                                                     & \multirow{2}{*}{0.192}                                        & \multirow{2}{*}{8}  & \multirow{2}{*}{90} & \multirow{2}{*}{5000}                                     \\ \cline{2-5}
                       & Test 2 & 0.0910                                                      & 0.0747                                                      & 0.1257                                                     &                                                               &                     &                     &                                                           \\ \hline
    \multirow{2}{*}{2} & Test 1 & 0.0674                                                      & 0.0685                                                      & 0.1910                                                     & \multirow{2}{*}{0.220}                                        & \multirow{2}{*}{8}  & \multirow{2}{*}{70} & \multirow{2}{*}{5000}                                     \\ \cline{2-5}
                       & Test 2 & 0.1245                                                      & 0.0962                                                      & 0.1210                                                     &                                                               &                     &                     &                                                           \\ \hline
    \multirow{2}{*}{3} & Test 1 & 0.0380                                                      & 0.0444                                                      & 0.1490                                                     & \multirow{2}{*}{0.165}                                        & \multirow{2}{*}{4}  & \multirow{2}{*}{90} & \multirow{2}{*}{5000}                                     \\ \cline{2-5}
                       & Test 2 & 0.0773                                                      & 0.0463                                                      & 0.1215                                                     &                                                               &                     &                     &                                                           \\ \hline
    \multirow{2}{*}{4} & Test 1 & 0.0731                                                      & 0.0658                                                      & 0.1919                                                     & \multirow{2}{*}{0.230}                                        & \multirow{2}{*}{16} & \multirow{2}{*}{70} & \multirow{2}{*}{5000}                                     \\ \cline{2-5}
                       & Test 2 & 0.0547                                                      & 0.0399                                                      & 0.1634                                                     &                                                               &                     &                     &                                                           \\ \hline
                       \hline
    \end{tabular}
    \end{minipage}
    \caption{\scriptsize Testing errors of four different LSTM networks for DTBM vessel in WMO sea state 8 at Froude number 0.4.}
\label{tab:architectures}
    \end{subfigure}
    
    \medskip
    \caption{\textit{LSTM test results (4 hidden layers, 90 neurons) network architecture 3, RSE=0.165}. {Heave, Pitch and Roll} motion dynamics of a notional DTBM battleship sailing in WMO sea state 8 at Froude number 0.4.  The inputs provided to the network are shown in \cref{fig:WaveCuts} (a),(b) (left column) and \cref{fig:WaveCuts} (c),(d) (right column) corresponding to 5000 train steps. Each time step is $\Delta t = 0.2s$.}
\label{fig:13}
\end{figure*}

In a similar fashion as in \cref{sec:Catmaran}, using deep recurrent neural networks with LSTM cells, we attempt to approximate the right functionals that, given the sea surface time series, will accurately estimate the motions of the DTBM vessels. A total of four examples (each with 3DOF) of the architecture convergence are shown in \cref{fig:13}. In it we can see two test sequences. The vessel motions for the two test cases are computed from unseen sea state realizations presented in \cref{fig:WaveCuts}. The motions predicted by the LSTM are heave, pitch and roll motions. We compare these to the motions predicted from the CFD code (for unused data) to calculate the relative squared error (RSE) for each set of architecture parameters;
the RSE is defined as follows:

\begin{equation}
RSE=\frac{\sum_{j=i}^{n}\left(Y_{i}-\hat{Y}_{i}\right)^{2}}{\sum_{j=1}^{n}\left(\hat{Y}_{i}-mean(\hat{Y}_{i})\right)^{2}}
.
\end{equation}

The architecture convergence shows consistent and accurate approximation of the vessel motions in the extreme sea states. The performance in roll motions is somewhat worse but this can be expected given the higher complexity of the dynamics. Fig. \ref{fig:13} shows results from the network architecture with the best performance. We see that roll motions with large amplitudes, almost 20 degrees, are accurately approximated both in amplitude and frequency of the time signal. The same can be said about the heave and pitch motions. The table in \cref{fig:13} summarizes the results
for all the LSTM architectures tested in our experiments.

The overall conclusions of the architecture convergence are similar to those obtained in \cref{sec:Catmaran}. Given more data we should be able to train more complex networks to improve the accuracy of the motion prediction. This data could potentially come from several different sources and possibly a real vessel, for which approximating a lifetime or near future motions is interesting. These will be vessels that may be required to operate in very harsh weather conditions. Good examples are military, coastguard or rescue vessels. 

%\begin{figure}[!h]
 %       \centering
 %       \includegraphics[width=0.8\columnwidth]{Arleigh_Burke.jpg}
 %       
 %       \medskip
 %
 %       \caption{\scriptsize USS Arleigh Burke (DDG-51) in rough seas, 31 Mar 1993.}
 %      \label{fig:Burke}
%\end{figure}

\section{Conclusions}

We have successfully trained deep recurrent neural networks of LSTM type to approximate nonlinear motions in irregular long crested head and oblique seas. To the authors' best knowledge, this constitutes a new paradigm in simulating the motion of marine vessels in mild or extreme sea states. A possible generalization that requires more extensive training is to train the LSTM network for different sea states and also different wave spectra associated with different geographic regions. This offline training is of course time consuming but the online predictions yields truly real-time predictions.  Given that the available wave radars and even satellite images that can provide the sea state accurately, pre-trained LSTM like the one we demonstrated here represent a new approach of simulating in a quick and accurate way vessel motions. Under extreme weather conditions this would constitute a potentially powerful predictive tool to avoid associated hazards encountered in these situations. 

This new approach is not limited to the specific application presented here but it can be applied to many other
nonlinear dynamical systems, where the aim is to forecast specific outputs. Moreover, an extension of the theorem of \cite{Functionals}
states that a neural network properly designed can also approximate continuous nonlinear operators \cite{chen1995universal} assuming that input functions belong to a compact set. The first such network, termed DeepONet, has been published in \cite{lu2019deeponet}, and can also be tested in the future for the application presented in the current work.

%%%%%%%%%% Insert bibliography here %%%%%%%%%%%%%%
\bibliographystyle{IEEEtran}
\bibliography{RSPA_Author_tex}

%%%%%%%%%%%%%%%%%%%%%%%%%%%%%%%%%%%%%%%%%%%%%%%%%%%%%%%%%%%%%%%%%%%%%%%%%%%%%%%%%%%%%%%%%%%%%%%%%
%%%%%%%%%%%%%%%%%%%%%%%%%%%%%%%%%%%%%%%%%%%%%%%%%%%%%%%%%%%%%%%%%%%%%%%%%%%%%%%%%%%%%%%%%%%%%%%%%
%%%%%%%%%%%%%%%%%%%%%%%%%%%%%%%%%%%%%%%%%%%%%%%%%%%%%%%%%%%%%%%%%%%%%%%%%%%%%%%%%%%%%%%%%%%%%%%%%

%%%%%%%%%%%%%%%%%%%%%%%%%%%%%%%%%%%%%%%%%%%%%%%%%%%%%%%%%%%%%%%%%%%%%%%%%%%%%%%%%%%%%%%%%%%%%%%%%
%%%%%%%%%%%%%%%%%%%%%%%%%%%%%%%%%%%%%%%%%%%%%%%%%%%%%%%%%%%%%%%%%%%%%%%%%%%%%%%%%%%%%%%%%%%%%%%%%
%%%%%%%%%%%%%%%%%%%%%%%%%%%%%%%%%%%%%%%%%%%%%%%%%%%%%%%%%%%%%%%%%%%%%%%%%%%%%%%%%%%%%%%%%%%%%%%%%

%%%%%%%%%%%%%%%%%%%%%%%%%%%%%%%%%%%%%%%%%%%%%%%%%%%%%%%%%%%%%%%%%%%%%%%%%%%%%%%%%%%%%%%%%%%%%%%%%%%%%%%%%%%%%%

\begin{table}[h!]

\end{table}

\end{document}